\documentclass[10pt,twocolumn,letterpaper]{article}

\usepackage[pagenumbers]{cvpr} % To force page numbers, e.g. for an arXiv version

\usepackage{graphicx}
\usepackage{amsmath}
\usepackage{amssymb}
\usepackage{booktabs}
\usepackage{svg}

\usepackage[pagebackref=true,breaklinks=true,colorlinks=true,bookmarks=false,linkcolor={red!50!black},urlcolor={darkgray!80!black},citecolor={green!50!black}]{hyperref} 

\usepackage[marginal]{footmisc}

\newcommand{\wlink}[1]{\textcolor{magenta}{{#1}}}

\usepackage[capitalize]{cleveref}
\crefname{section}{Sec.}{Secs.}
\Crefname{section}{Section}{Sections}
\Crefname{table}{Table}{Tables}
\crefname{table}{Tab.}{Tabs.}

\usepackage{xcolor}
\usepackage{enumitem}
\usepackage{xspace}
\usepackage{booktabs}
\usepackage{colortbl}
\usepackage{multirow}
\usepackage{makecell}
\usepackage{lipsum} % Just for generating dummy text, can be removed
\usepackage{gensymb} % for \degree

\newcommand{\Tref}[1]{Table~\ref{#1}}
\newcommand{\eref}[1]{Eq.~\eqref{#1}}

\newcommand{\fref}[1]{Fig.~\ref{#1}}
\newcommand{\Fref}[1]{Figure~\ref{#1}}

\newcommand{\Sref}[1]{Section~\ref{#1}}

\usepackage[labelsep=period]{caption}
\renewcommand{\paragraph}[1]{\vspace{0.2em}\noindent \textbf{#1 \hspace{0.2em}}}

\definecolor{MyDarkRed}{rgb}{0.66, 0.16, 0.16}
\definecolor{MyDarkBlue}{rgb}{0.16, 0.16, 0.66}

\def\paperID{3586} % *** Enter the CVPR Paper ID here
\def\confName{CVPR}
\def\confYear{2024}

\begin{document}

\newcommand{\MethodName}{RichDreamer\xspace}
\title{\MethodName: A Generalizable Normal-Depth Diffusion Model for \\ Detail Richness in Text-to-3D}

\author{Lingteng Qiu$^{1,3}\footnotemark[1]~\footnotemark[2]$ \quad Guanying Chen$^{2,1}\footnotemark[1]$ \quad Xiaodong Gu$^{3}\footnotemark[1]$ \\ Qi Zuo$^{3}$ \quad Mutian Xu$^{1}$ \quad Yushuang Wu$^{2,1,3}$ \quad Weihao Yuan$^{3}$ \quad Zilong Dong$^{3}$ \\ Liefeng Bo$^{3}$, \quad Xiaoguang Han$^{1,2}\footnotemark[3]$ \vspace{0.3em} \\
{\normalsize $^1$SSE, CUHKSZ}
\quad{\normalsize $^2$FNii, CUHKSZ} \quad {\normalsize $^3$Alibaba Group}
}

\twocolumn[{
\renewcommand\twocolumn[1][]{#1}
\maketitle

\vspace{-18pt}
\begin{center}
    \captionsetup{type=figure}
    \includegraphics[width=\textwidth]{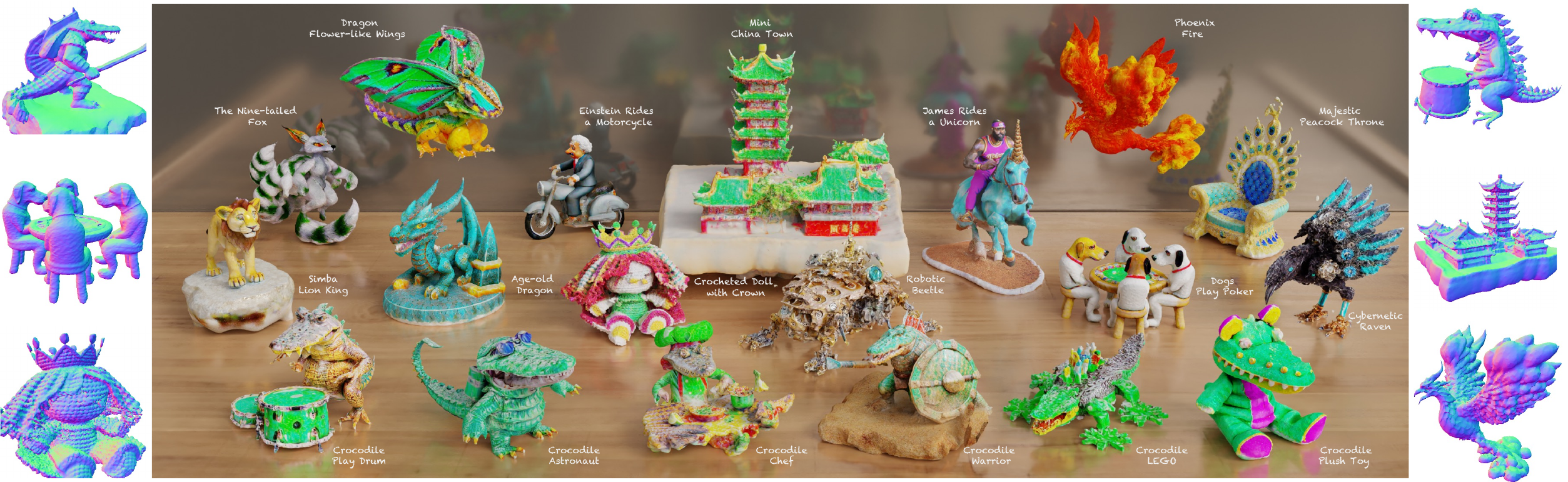}

    \captionof{figure}{\textbf{3D Generation Results and Applications of \emph{\MethodName}}. \MethodName can generate highly-detailed and diverse 3D content from free-form user prompts. 
    Our method achieves this by first generating the object geometry based on a generalizable Normal-Depth diffusion model, followed by modeling the physically-based rendering (PBR) materials.
    Notably, the diverse crocodile-theme objects at the bottom highlights the generalization ability of our method.
    The abbreviation of text prompts are shown beside the corresponding objects (full prompts can be found in the supplementary materials).
    }
    
    \label{fig:teaser}
\end{center}
}]

\footnotetext[1]{Equal contribution.}
\footnotetext[2]{Work done during internship at Alibaba.}
\footnotetext[3]{Corresponding author:
\wlink{hanxiaoguang@cuhk.edu.cn}.}

%%%%%%%%% ABSTRACT
\begin{abstract}
    Lifting 2D diffusion for 3D generation is a challenging problem due to the lack of geometric prior and the complex entanglement of materials and lighting in natural images.
    Existing methods have shown promise by first creating the geometry through score-distillation sampling (SDS) applied to rendered surface normals, followed by appearance modeling. 
    However, relying on a 2D RGB diffusion model to optimize surface normals is suboptimal due to the distribution discrepancy between natural images and normals maps, leading to instability in optimization.
    In this paper, recognizing that the normal and depth information effectively describe scene geometry and be automatically estimated from images, we propose to learn a generalizable Normal-Depth diffusion model for 3D generation. We achieve this by training on the large-scale LAION dataset together with the generalizable image-to-depth and normal prior models. 
    %We achieve this by training on the large-scale LAION dataset and fine-tuning on the synthetic Objaverse dataset.
    In an attempt to alleviate the mixed illumination effects in the generated materials, we introduce an albedo diffusion model to impose data-driven constraints on the albedo component.
    Our experiments show that when integrated into existing text-to-3D pipelines, our models significantly enhance the detail richness, achieving state-of-the-art results. Our project page is \wlink{https://aigc3d.github.io/richdreamer/}.
\end{abstract}

%    In addition, to explore potential improvements in addressing mixed illumination effects in the generated materials, we introduce an albedo diffusion model to impose data-driven constraints on the albedo component.
%    In addition, to alleviate the mixed illumination effects in the generated materials, we introduce an albedo diffusion model to impose data-driven constraints on the albedo component \todo{Weaken albedo}. Our experiments show that integrating our models into existing text-to-3D pipelines yields state-of-the-art results in both geometry and appearance modeling. 

\section{Introduction}
\label{sec:intro}
% Background: progress in 2D generation, 3D generation, and applications
Image generation models have witnessed notable advancements in controllable image synthesis \cite{ramesh2021zero,rombach2022high}. This remarkable progress can be attributed to the scalability of generative models \cite{ho2020denoising,song2020denoising} and utilization of the large-scale training datasets consisting of billions of image-caption pairs scrapped from the internet \cite{schuhmann2022laion}.
Conversely, due to the limited scale of the publicly available 3D datasets, existing 3D generative models are primarily evaluated for category-specific generation and face challenges when attempting to generate novel, unseen categories \cite{gao2022get3d,gupta20233dgen,zhang20233dshape2vecset}. 
It remains an open problem to create a comprehensive 3D dataset to facilitate generalizable 3D generation.

% Existing methods and problems
Recently developments in the field of text-to-3D, such as DreamFusion \cite{poole2022dreamfusion}, have demonstrated impressive capabilities in zero-shot generation. This is achieved by optimizing a neural radiance field \cite{mildenhall2020nerf} through score distillation sampling (SDS) \cite{poole2022dreamfusion,wang2023score} using a 2D diffusion model \cite{saharia2022photorealistic}.
Subsequent to this,  several methods have been proposed to improve the quality of the generated objects \cite{lin2023magic3d,wang2023prolificdreamer,chen2023fantasia3d,metzer2023latent}.
However, the approach of lifting from 2D to 3D has presented two primary challenges.
Firstly, 2D diffusion models tend to lack multi-view constraints, often leading to the emergence of multi-face objects, a phenomenon referred to as the ``Janus problem'' \cite{poole2022dreamfusion,wang2023score}.
To address this issue, recent advancements in multi-view-based diffusion models have shown success in mitigating these multi-face artifacts \cite{shi2023mvdream,zhao2023efficientdreamer}.
%(\eg, MVDream \cite{shi2023mvdream} and EfficientDreamer \cite{zhao2023efficientdreamer}) 

Secondly, given the inherent coupling of surface geometry, texture, and lighting in natural images, the direct use of 2D diffusion models for the simultaneous inference of geometry and texture is considered suboptimal \cite{chen2023fantasia3d}.
This suggests a two-stage decoupled approach: first, the generation of geometry, followed by the generation of texture.
The recent Fantasia3D \cite{chen2023fantasia3d} method has shown promise in this decoupling strategy, yielding notably improved geometric reconstructions.
However, Fantasia3D relies on 2D RGB diffusion models to optimize normal maps, leading to data distribution discrepancies that compromise the quality of geometric generation and introduce instability in optimization.
This limitation underscores the pressing need for a robust prior model to provide an effective geometric foundation.

% Geometry
In this work, we aim to develop a robust 3D prior model to push forward the decoupled text-to-3D generation approach. 
Creating a model that offers 3D geometric priors typically requires access to 3D data for training supervision.
However, amassing a large-scale dataset containing high-quality 3D models across diverse scenes is a challenging endeavor due to the time-consuming and costly process of 3D object scanning and model design \cite{reizenstein2021common,wu2023omniobject3d,yu2023mvimgnet,deitke2023objaverse}. 
The limited scale of the publicly available 3D datasets presents a critical challenge: \emph{how to learn a generalizable 3D prior model with limited 3D data}?

Recognizing that the normal and depth information can effectively describe scene geometry and can be automatically estimated from images \cite{ranftl2021vision}, we propose to learn a generalizable \emph{Normal-Depth diffusion model} for 3D generation (see \fref{fig:teaser}). 
This is achieved by training on the large-scale LAION dataset \cite{schuhmann2022laion} to learn diverse distributions of normal and depth of real-world scenes with the help of the generalizable image-to-depth and normal prior models \cite{ranftl2020towards,bae2021estimating}. 
To improve the capability in modeling a more accurate and sharp distribution of normal and depth,  
we fine-tune the proposed diffusion model on the synthetic Objaverse dataset \cite{deitke2023objaverse}. 
Our results demonstrate that by pre-training on a large-scale real world dataset, the proposed Normal-Depth diffusion model can retain its generalization ability after fine-tuning on the synthetic dataset, indicating that our model learns a good distribution of diverse normal and depth in real-world scenes. 
%with our results underscoring the significance of training on a large-scale real-world dataset to retain generalization ability.
%with the help of the large-scale LAION dataset together with the generalizable image-to-depth and normal pretrained models, %

% Textures
Given the inherent ambiguity in the decomposition of surface reflectance and illumination effects, textures generated by existing methods often retain shadows and specular highlights \cite{richardson2023texture}. 
In an attempt to address this problem, we introduce an \emph{albedo diffusion model} to provide data-driven constraints on the albedo component, enhancing the disentanglement of reflectance and illumination effects.

In summary, the key contributions of this paper are as follows:  
\begin{itemize}[itemsep=0pt,parsep=0pt,topsep=2bp]
    \item We propose a novel Normal-Depth diffusion model to provide strong geometric prior for high-fidelity text-to-3D geometry generation. By training on the extensive LAION dataset, our method exhibits remarkable generalization abilities.
    \item We introduce an albedo diffusion model that acts as a data-driven regularization for albedo, resulting in a more accurate separation of reflectance and illumination effects.
    \item Experiments demonstrate that integrating our models into existing text-to-3D pipelines yields state-of-the-art results in both geometry and appearance modeling. 

\end{itemize}

\begin{figure*}[tb] \centering
    \includegraphics[width=\textwidth]{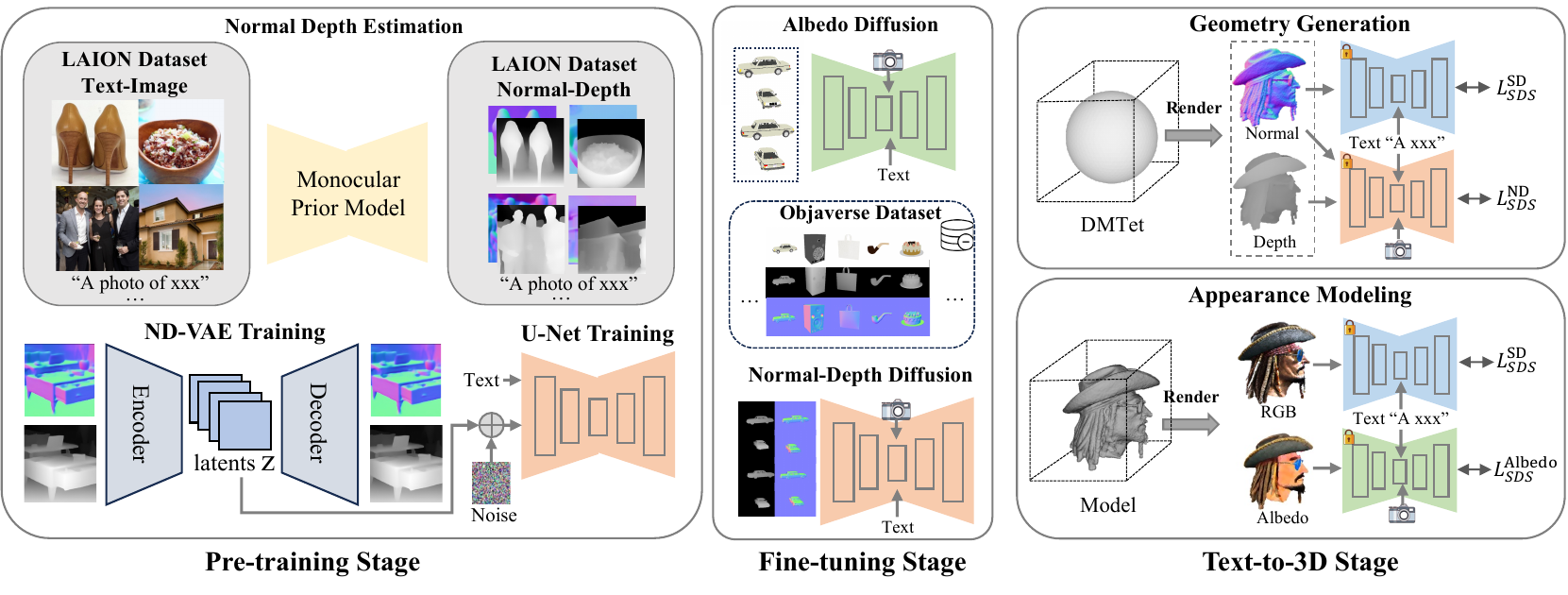}
    \caption{\textbf{Overview of the proposed \emph{\MethodName}}. 
    We introduce a generalizable Normal-Depth diffusion model that is trained on the LAION-2B dataset with normal and depth predicted by Midas \cite{ranftl2020towards}, followed by fine-tuning on the synthetic dataset.
    Our model can be incorporated with the DMTet and NeRF representations to enhance the geometry generation.
    To alleviate the ambiguity in appearance modeling, we propose an albedo diffusion model to impose data-drive prior on the albedo component.} 
    \label{fig:pipeline}
    \vspace{-3mm}
\end{figure*}

%-------------------------------------------------------------------------
\section{Related Work}
\label{sec:related_works}
\paragraph{3D Generative Model}
The creation of high-quality 3D content is gaining increasing importance in various applications.
Generative models directly learn the data distributions of the 3D data, enabling data sampling. Existing methods have yielded promising results by representing scenes using various 3D representations, including voxels \cite{wu2016learning,henzler2019escaping}, point clouds~\cite{nichol2022point, zeng2022lion, luo2021diffusion}, meshes~\cite{liu2023meshdiffusion,gao2022get3d}, and implicit fields~\cite{wang2023rodin,cheng2023sdfusion,karnewar2023holofusion, anciukevivcius2023renderdiffusion, muller2023diffrf, jun2023shap, zhang20233dshape2vecset, gupta20233dgen, erkocc2023hyperdiffusion, chen2023single,yu2023pushing}.
However, these methods primarily demonstrate their generative capabilities within limited categories of objects due to the restricted scale of 3D training datasets. In contrast, our approach addresses the text-to-3D problem by extending 2D diffusion to the 3D domain, allowing for better generalization across diverse scenes specified by user prompts.

\paragraph{2D Diffusion for 3D Generation}
Recent research has demonstrated the generation of 3D objects from user prompts, leveraging pre-trained models like CLIP model \cite{radford2021learning,jain2022zero,xu2023dream3d,mohammad2022clip} or 2D diffusion models \cite{saharia2022photorealistic,rombach2022high}.
Notably, DreamFusion \cite{poole2022dreamfusion} achieves zero-shot text-to-3D generation by optimizing a neural radiance field (NeRF) \cite{mildenhall2020nerf} through score distillation sampling (SDS) with a 2D diffusion model. Concurrently, SJC \cite{wang2023score} employs score Jacobian chaining for the 3D generation.

Encouraged by these promising results, numerous works have focused on improving the quality of generated objects through approaches such as coarse-to-fine optimization \cite{lin2023magic3d,cheng2023progressive3d}, decoupled generation \cite{chen2023fantasia3d}, new score distillation \cite{wang2023prolificdreamer}, improved optimization strategies \cite{wu2023hd,chen2023it3d,metzer2023latent,tsalicoglou2023textmesh,huang2023dreamtime,armandpour2023re,hong2023debiasing,armandpour2023re,xu2023matlaber,seo2023let}, and incorporating parametric shape model \cite{huang2023dreamwaltz,han2023headsculpt,liao2023tada,zhang2023dreamface} etc.
As generating a 3D model typically involves hours of optimization, some methods explore efficient 3D representations (\eg, hashgrid \cite{muller2022instant} and 3D Gaussian splatting \cite{kerbl20233d}) or improved training strategy to accelerate the optimization \cite{yi2023gaussiandreamer,tang2023dreamgaussian,lorraine2023att3d,threestudio2023}.
Alongside the rapid development of the text-to-3D field, several methods have also adopted 2D diffusion models for the problem of 3D generation from image condition \cite{zhou2023sparsefusion,gu2023nerfdiff,deng2023nerdi,melas2023realfusion,xu2022neurallift,tang2023make,qian2023magic123,liu2023one,raj2023dreambooth3d,anciukevivcius2023renderdiffusion,zhao2023michelangelo}.
However, as 2D diffusion models tend to lack multi-view constraints, these methods often suffer from the multi-view inconsistency issue, resulting in less desirable 3D generation outcomes.

\paragraph{Geometry Prior for Diffusion Models}
To enhance diffusion models for generative novel-view synthesis \cite{liu2023zero,watson2022novel,chan2023generative}, Zero-1-to-3 \cite{liu2023zero} fine-tunes the Stable Diffusion model \cite{rombach2022high} to synthesize novel views conditioned on relative poses.
Recent approaches have significantly improved the consistency of the generated multi-view images by performing multi-view diffusion \cite{shi2023mvdream,liu2023syncdreamer,ye2023consistent,weng2023consistent123,shi2023zero123++,tseng2023consistent,zhao2023efficientdreamer,szymanowicz2023viewset}.
For example, MVDream \cite{shi2023mvdream} fine-tunes a pre-trained diffusion model on the synthetic Objaverse dataset \cite{deitke2023objaverse} to simultaneously generate a set of 4-view images of the same object, conditioned on the camera poses. 
While these methods effectively address the multi-view inconsistency problem, they perform diffusion in RGB image space, making them less suitable for the decoupled generation approach where the geometry is generated before appearance.

There are methods incorporating more explicit geometric constraints into the diffusion models. %\cite{long2023wonder3d,huang2023humannorm,li2023sweetdreamer}. 
LDM3D \cite{stan2023ldm3d} introduces an RGB-D diffusion model on the LAION-400M dataset. However, this model is not validated for text-to-3D generation.
Concurrent to our work, SweetDreamer \cite{li2023sweetdreamer} proposes to align the geometric prior in 2D diffusion using a canonical coordinate map (CCM) representation. However, CCM implicitly requires the training objects to be aligned and can only be obtained from synthetic 3D datasets, potentially limiting its generalization and scalability.
Wonder3D \cite{long2023wonder3d} introduces an RGB-Normal diffusion model on the Objaverse dataset.
HumanNorm \cite{huang2023humannorm} proposes two disjoint diffusion models, one for normal and the other for depth, on a 3D Human dataset of 2952 body models. 
In contrast, our model jointly learns the distribution of normal and depth, and was pre-trained on the extensive LAION-2B dataset to improve the generalization ability. In addition, we introduce an albedo diffusion model to better model appearance.

% text-to3d based on Clip 
%Some methods investigate better strategy to stabilize the optimization process, including sampling time scheduling and score d  \cite{seo2023let,tsalicoglou2023textmesh,huang2023dreamtime,armandpour2023re,hong2023debiasing,armandpour2023re}. 
%Early approaches adopts the \cite{jain2022zero,xu2023dream3d,mohammad2022clip}
% text-to-3d with diffusion
%Multi face problems
%As these methods directly use
%Our method focus on text-to-3D, Multiface problems\cite{}
%These method relies on 2D RGB diffusion models for 3D generation, lacks.
% image to 3d 
%\paragraph{3D Representation}
%
%\paragraph{image-to-3d Generation}
%\subsection{Text-to-image Generation}
%Recently, 2D Diffusion models have made significant progress in generating photorealistic images from text prompts~\cite{ho2020denoising,ramesh2021zero,rombach2022high,gal2022image,yang2022diffusion,ramesh2022hierarchical,saharia2022photorealistic,croitoru2023diffusion}.
% refer to Magic3d, HD-Fusion
%\paragraph{Diffusion Model for Novel-view Synthesis}
% novel-view synthesis
% normal diffusion

\newcommand{\point}{\mathbf{x}}
\newcommand{\dir}{\mathbf{d}}
\newcommand{\pointcolor}{\mathbf{c}}
\newcommand{\density}{\sigma}
\newcommand{\raycolor}{\hat{\mathbf{C}}}
\newcommand{\ray}{\mathbf{r}}
\newcommand{\nerffun}{\mathbf{f}}
\newcommand{\nerfparam}{\phi}

\newcommand{\data}{x}
\newcommand{\latent}{z}
\newcommand{\noise}{\epsilon}
\newcommand{\dmparam}{\theta}
\newcommand{\albedo}{a}
\newcommand{\dmparamalbedo}{\theta_\albedo}
\newcommand{\camera}{c}
\newcommand{\depth}{d}
\newcommand{\dmtime}{t}
\newcommand{\image}{x}
\newcommand{\volumerender}{g}
\newcommand{\prompt}{y}
\newcommand{\encoder}{\mathcal{E}}

\newcommand{\real}{\mathbb{R}}
\newcommand{\diffuse}{k_d}
\newcommand{\roughness}{k_r}
\newcommand{\metallic}{k_m}
\newcommand{\ntangent}{k_n}
\newcommand{\mlpmaterial}{f_\parammat}
\newcommand{\parammat}{\psi}

\section{Method}
%In this section, we detail our approach for text-to-3D generation. 
%Our approach for text-to-3D generation follows a two-stage decoupled generation strategy, where we first generate the geometry of the 3D content, followed by the stage of appearance modeling.
In this section, we introduce a Normal-Depth diffusion model and an albedo diffusion model to push forward the decoupled 3D generation pipeline, where the geometry is first generated followed by the appearance modeling (see \fref{fig:pipeline}).

\paragraph{Overview}
The existing approach for decoupled generation \cite{chen2023fantasia3d} adopts a text-to-image diffusion model (\ie, Stable-Diffusion \cite{rombach2022high}) to optimize the rendered surface normals of the object for geometry generation. 
However, this direct application is suboptimal due to the discrepancy in data distribution between natural images and normal maps, often leading to unstable optimization and compromised geometric fidelity \cite{chen2023fantasia3d}. 
As a result, appropriate geometry initialization (\eg, 3D ellipsoids with different shapes and orientations, or coarse 3D models) is often needed to achieve good results for different prompts.
In response, we propose to learn a 3D-aware diffusion model tailored for 3D geometry generation. 
Specifically, we introduce a Normal-Depth diffusion model pre-trained on an extensive real-world dataset and further fine-tuned on a synthetic dataset, to offer consistent guidance for geometry generation.

Another critical issue in text-to-3D generation is the inaccurate appearance modeling, where materials intermingle with the lighting effects like shadows and specular highlights, often resulting in imprecise relighting. 
In an attempt to address this, our method integrates an albedo diffusion model to regularize the albedo component of the materials, effectively separating the albedo from the influence of lighting artifacts.

\subsection{Normal-Depth Diffusion Model}
To endow the diffusion model with 3D geometric priors for 3D generation tasks, we introduce a novel Normal-Depth diffusion model.
Different from existing methods that either learn a normal or a depth diffusion model \cite{huang2023humannorm,stan2023ldm3d}, our model captures the joint distribution of normal and depth, leveraging their intrinsic complementary nature-depths describe the macrostructure of the scene while normals provide local surface details. 

\paragraph{Model Architecture}
We adapted the architecture of the publicly available text-to-image diffusion model, Stable Diffusion (SD) \cite{rombach2022high}, with minor modifications.
SD incorporates a variational auto-encoder (VAE) with KL-regularization \cite{kingma2013auto}, and a latent diffusion model (LDM). 
The VAE maps an image of size $512\times 512$ to and from a latent space of size $64\times 64$, and the LDM is a UNet denoiser that learns to denoise the latent feature guided by the text prompt.

For our purpose, we extended the input and output channel number of SD's VAE from three to four channels to encompass three for normals and one for depth, keeping other components unchanged. 

\paragraph{Pre-training on Real-world Data}
The LAION-2B dataset, comprising billions of correlated image and text pairs, served as our foundational training resource \cite{schuhmann2022laion}. 
We prepared our training set with text prompts paired with corresponding normal and depth maps, utilizing NormalBae~\cite{bae2021estimating} and Midas-3.1~\cite{ranftl2020towards}, which are leading methods for monocular normal and depth estimation.

We first trained the Normal-Depth VAE to learn the joint distribution of normal and depth with the MSE reconstruction loss, adversarial loss, and the KL-regularization loss \cite{rombach2022high,esser2021taming}.
We then trained the LDM to enable text to Normal-Depth generation. Denoting $\data$ as the normal and depth data, $\encoder$ the encoder, $z$ the latent feature, and $\noise_{\dmparam}$ the UNet denoiser, the objective for training LDM can be written as:
\begin{align}
    \mathcal{L}_{\text{LDM}} = \mathbb{E}_{\latent\sim \mathcal{E}(\data), \prompt, \dmtime, \noise \sim \mathcal{N}(0,1)}\left[\|\noise_{\dmparam}(\latent_\dmtime, \prompt, \dmtime) - \noise\|^2_2\right],
\end{align}
where $\latent_\dmtime$ is the noised latent variable at a specific denoising timestep $\dmtime$, 
and $\prompt$ the text embedding obtained from the CLIP model \cite{radford2021learning}.
Our results show that the pre-training on the real-world dataset is crucial to maintain the generalization ability on diverse prompts.

\paragraph{Fine-tuning on Synthetic 3D Data}
To enhance object-level 3D generation, we fine-tuned our Normal-Depth LDM on the Objaverse dataset \cite{deitke2023objaverse}, which features ground-truth 3D models. 
We render the ground-truth normal and depth maps with the provided object. 
In the fine-tuning stage, we employed a four-view diffusion technique proposed by MVDream \cite{shi2023mvdream}. The camera poses are mapped by a simple Multilayer Perceptron (MLP) to be the camera embeddings, which will be added to the time embedding to be accessed by the diffusion model.
The training objective in the fine-tuning stage is:
\begin{align}
    \mathcal{L}_{\text{LDM}}^{'} = \mathbb{E}_{\latent, \prompt, \camera, \dmtime, \noise \sim \mathcal{N}(0,1)}\left[\|\noise_{\dmparam}(\latent_\dmtime, \prompt, \camera, \dmtime) - \noise\|^2_2\right],
\end{align}
where $c$ is the camera condition.

\paragraph{Implementation Details}
For training on the LAION dataset, we follow the data filtering strategy used in the training of SD v2.1 to ensure high-quality data selection. 
The image is resized to $384\times 384$ as input to Midas for estimating normal and depth. 
The computational expense for training the Normal-Depth VAE and LDM amounted to $1,344$ and $11,520$ GPU hours respectively on A100-80G GPUs. More details can be found in the supplementary materials

For fine-tuning on the Objaverse dataset, for each 3D object, we established camera positions within a radial distance of 1.4 to 2.0 units and an elevation angle spanning 5 to 30 degrees. We rendered 24 views per object, distributed uniformly across azimuth angles. To enhance the dataset quality, we discarded objects whose rendered images scored low in relevance to the object names as determined by CLIP scores, resulting in a pool of 270,000 training objects. The text prompts used in training were obtained by a hybrid approach: 30\% stemmed from object tags and names, and the remaining 70\% were from Cap3D \cite{luo2023scalable}.
The fine-tuning utilized a batch size of $512$ with gradient accumulation performed every $8$ batch.
This was conducted on 8 GPUs for one week, reaching a total of $20,000$ iterations.

\subsection{Geometry Generation}
\label{sec:geometry}
\paragraph{Score Distillation Sampling (SDS)}
Existing 2D lifting approaches for text-to-3D typically employ either a NeRF representation \cite{mildenhall2020nerf,poole2022dreamfusion} or the hybrid DMTet representation \cite{shen2021deep,lin2023magic3d} to represent the 3D content.
Denoting $\nerfparam$ as the parameters of a 3D representation and $\volumerender$ as the differentiable rendering function, the rendered image can be expressed as $\image=\volumerender(\nerfparam)$.
DreamFusion \cite{poole2022dreamfusion} introduces a Score Distillation Sampling (SDS) process that leverages gradient-based score functions to guide the optimization of parameters in 3D representation for object generation:
\begin{equation}
    \label{eq:sds}
    \nabla_{\nerfparam} \mathcal{L}_{\text{SDS}}(\phi, \image=\volumerender(\nerfparam)) = \mathbb{E}_{\dmtime, \noise}\left[w(t)(\noise_{\dmparam}(\latent_\dmtime ; \prompt, \dmtime) - \noise) \frac{\partial \image}{\partial \nerfparam}\right],
\end{equation}
where the expression $(\noise_{\dmparam}(\latent_t; \prompt, t) - \noise)$ represents the difference between the actual noise $\noise$ and the noise estimated by the UNet $\noise_{\dmparam}$. The term $w(t)$ is a weighting term that depends on the timestep $t$, and $\prompt$ indicates the text embedding.

Fantasia3D \cite{chen2023fantasia3d} shows that the SDS loss derived from the image diffusion model (\eg, Stable Diffusion \cite{rombach2022high}) can be applied to the rendered normal maps from a DMTet for geometry generation.

\paragraph{Normal-Depth Diffusion for 3D Generation}
Compared to the image diffusion model, our Normal-Depth diffusion model is specifically designed for modeling the joint distribution of normal and depth maps. It provides effective supervision for geometry optimization.

We utilize a DMTet representation and integrate our Normal-Depth diffusion model into the coarse-to-fine geometry generation pipeline of Fantasia3D \cite{chen2023fantasia3d}. The normal and depth of the object can be efficiently rendered using differentiable rasterization \cite{laine2020modular}. The geometry generation loss function is defined as:
\begin{align}
    \label{eq:sds_geo}
    \mathcal{L}_\text{Geo} = \lambda_\text{SD} \mathcal{L}_{\text{SDS}-\text{Normal}}^\text{SD} + \lambda_\text{ND} \mathcal{L}_{\text{SDS}-\text{ND}}^\text{ND},
\end{align}
where the first loss term is employed in Fantasia3D \cite{chen2023fantasia3d} to enforce SDS on the rendered normal maps using Stable Diffusion. The second loss term is enabled by our Normal-Depth diffusion model to impose SDS on the composite of the rendered normal and depth maps. By default, we initialize the DMTet as a Sphere.

\paragraph{Integration with NeRF}
Since normal and depth maps can be derived from the NeRF representation using volume rendering, our Normal-Depth diffusion model can also be utilized to optimize NeRF using the loss defined in \eref{eq:sds_geo}. Given that the normal and depth maps derived from NeRF can be noisy at the start of the optimization, we impose SDS loss on the rendered RGB images with SD during the first 1,000 iterations to warm up the optimization.

Recognizing that the NeRF representation is more flexible in modeling complex structures during optimization, we also investigate the idea of converting the optimized NeRF to DMTet as the initialization for geometry refinement.

\paragraph{Optimization}
For geometry generation, we accelerate the SDF function in DMTet with an efficient hash encoding~\cite{muller2022instant}.
The loss weights $\lambda_\text{SD}$ and $\lambda_\text{ND}$ are both set to $1$. The optimization process takes approximately $1.5$ hours on a single GPU
with $30$ GB of memory.

\subsection{Appearance Modeling}
%The second step in our pipeline involves the synthesis of appearance.

\paragraph{Physically-based Rendering}
For DMTet representation, in line with prior studies \cite{chen2023fantasia3d,Munkberg_2022_CVPR}, we employ the Physically Based Rendering (PBR) Disney material model \cite{burley2012physically}, which integrates a diffuse term with a specular GGX lobe \cite{walter2007microfacet}.
The material property of a surface point is determined by the diffuse color $\diffuse \in \real^3$, roughness $\roughness \in \real$, metallic term $\metallic \in \real$, and normal variation in tangent space $\ntangent \in \real^3$. 
We parameterize the spatially-varying materials of the surface by a learnable MLP $\mlpmaterial$ with parameters $\parammat$ to predict material parameters for input 3D point $p$ as:
\begin{align}
    \label{eq:}
    (\diffuse, \roughness, \metallic, \ntangent) = \mlpmaterial (p).
\end{align}
By specifying the environment lighting and the camera viewpoint, the image color can be computed using a differentiable renderer based on the surface geometry and materials \cite{Munkberg_2022_CVPR}.

The existing method optimizes materials by imposing SDS loss on the final rendered RGB images \cite{chen2023fantasia3d}. However, this approach may lead to inaccuracies in material decomposition due to the inherent challenges in disentangling material components based solely on color.

To regularize the material generation, an ideal prior model should effectively regularize both the diffuse and specular components.
However, due to the varied creation methods of existing 3D models and the lack of standardization \cite{deitke2023objaverse}, it is challenging to acquire a comprehensive dataset with consistent and accurate ground truth for the specular component.
In light of this difficulty, we introduce an albedo diffusion model to decouple the albedo from complex lighting effects, serving as a preliminary approach to mitigate the challenge of mixed illumination.

\paragraph{Depth-Conditioned Albedo Generation}
A direct solution for the albedo diffusion model involves fine-tuning a text-to-image diffusion model using paired data of text prompts and albedo maps. 
While this method is effective for sampling, it falls short for 3D generation due to potential misalignments between the generated albedo and the geometry.
To ensure the alignment of generated albedo maps with geometry, we employ the depth map from the corresponding viewpoint as a condition within the albedo Latent Diffusion Model (LDM).
Specifically, we concatenate the depth map with the latent features to serve as input for the UNet denoiser \cite{ramesh2021zero}.
We also employed the four-view diffusion strategy proposed by MV-Dream for the albedo diffusion model.
We fine-tune the SD 2.1 on the Objaverse dataset \cite{deitke2023objaverse} to capture the albedo distribution with the following training objective:
\begin{align}
    \mathcal{L}_{\text{Albedo}} = \mathbb{E}_{\latent^\albedo, \prompt, \camera, \dmtime, \noise \sim \mathcal{N}(0,1)}\left[\|\noise_{\dmparamalbedo}(\latent^\albedo_\dmtime, \prompt, \camera, \depth, \dmtime) - \noise\|^2_2\right],
\end{align}
where $\latent^\albedo$ represents the latent feature of the albedo map, and $\depth$ is the depth condition.

\begin{figure*}[tb] \centering
  
   \includegraphics[width=\textwidth,height=0.83\textwidth]{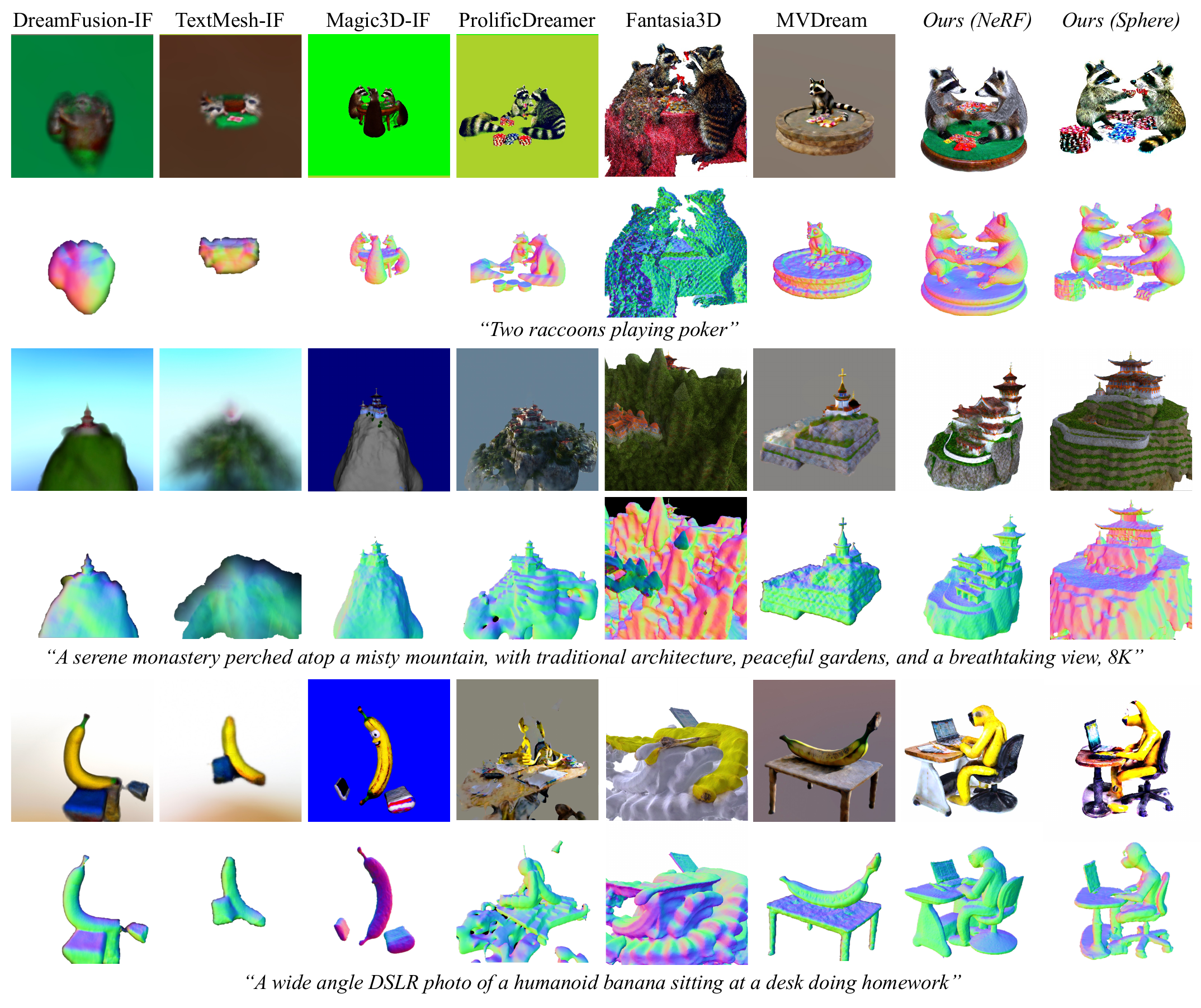}
   
    \vspace{-2mm}

    \caption{Visual comparison between our method and existing methods.}
    \label{fig:res_visual}
\end{figure*}

% \vspace{-5mm}
    
\paragraph{Loss Function}
The loss function for appearance modeling is expressed as:
\begin{align}
    \label{eq:sds_app}
    \mathcal{L}_\text{App} = \lambda_\text{SD} \mathcal{L}_{\text{SDS}-\text{RGB}}^\text{SD} + \lambda_\text{Albedo} \mathcal{L}_{\text{SDS}-\text{Albedo}}^\text{Albedo},
\end{align}
where the first term reflects the SDS on the rendered RGB images using SD, and the latter term is the SDS imposed on the albedo component by our Albedo diffusion model.

\paragraph{Optimization}
For appearance modeling, the loss weights $\lambda_\text{SD}'$ and $\lambda_\text{ND}'$ are also set to $1$. The optimization takes around $20$ minutes on a GPU.
After optimization, the material properties can be sampled at surface points and compiled into a 2D texture map \cite{chen2023fantasia3d,Munkberg_2022_CVPR,xatlas}, which can be directly imported into existing graphics engines for applications.

\begin{table*} \centering
    \vspace{-3mm}
    \caption{Quantitative comparison with existing methods. The geometry CLIP score is measured on the shading images rendered with uniform albedo, and the appearance CLIP score is measured on the images rendered with textured models (values the higher the better).}
    \resizebox{\textwidth}{!}{
    \begin{tabular}{r *{10}{c}}
        \toprule
        & DreamFusion-IF  & Magic3D-IF & TextMesh-IF & ProlificDreamer & Fantasia3D~(Sphere) & MVDream & \emph{Ours~(NeRF)} & \emph{Ours~(Shpere)}\\
        \midrule
        Geometry CLIP Score & 17.4548 & 20.1157 & 18.2222 & 23.3818 & 17.5398 & 24.8003 & 26.0570 & 25.8820 \\
        Appearance CLIP Score & 24.1091 & 27.8231 & 25.1218 & 31.8022 & 26.4055 & 28.7331 & 31.3551 & 31.7099\\
        \bottomrule
    \end{tabular}
    }
    \label{tab:res_quant}
    \vspace{-5mm}
\end{table*}

\section{Experiments}
\label{sec:Experiments}
In this section, we thoroughly evaluate the effectiveness of our proposed text-to-3D method by conducting a comprehensive comparison with state-of-the-art approaches.

\paragraph{Model Variants}
As discussed in \Sref{sec:geometry}, our Normal-Depth diffusion model can be applied to optimize DMTet and NeRF. 
To better verify the effectiveness of our Normal-Depth Diffusion model, we have designed two model variants for text-to-3D. 
The first model, denoted as \emph{Ours~(Sphere)}, initializes the DMTet with a Sphere for geometry generation. The second model, denoted as \emph{Ours~(NeRF)}, first optimizes a NeRF with our Normal-Depth diffusion model and then converts the NeRF to DMTet as an initialization for geometry generation.

\paragraph{Baselines}
We conducted extensive comparisons with a variety of baseline methods, including both DMTet-based and NeRF-based methods.
For the DMTet-based methods, we compared our approach against the state-of-the-art Fantasia3D method \cite{chen2023fantasia3d}, utilizing its publicly available official code with DMTet initialized as a Sphere.
In the case of NeRF-based methods, we evaluated our approach against multiple competitors, including DreamFusion-IF \cite{poole2022dreamfusion}, Magic3D-IF \cite{lin2023magic3d}, TextMesh-IF \cite{tsalicoglou2023textmesh}, and ProlificDreamer \cite{wang2023prolificdreamer}.
As there is no publicly available code for these four methods, we used the implementation from threestudio \cite{threestudio2023}, and IF indicates the DeepFloyd IF\footnote{\url{https://github.com/deep-floyd/IF}} diffusion model.
We also compared our method with the state-of-the-art NeRF-based method, MVDream \cite{shi2023mvdream}, using its publicly available official code.

SweetDreamer \cite{li2023sweetdreamer} is a contemporaneous work that is compatible with DMTet and NeRF. Given the absence of a public implementation, we conducted a fair comparison by evaluating our results against those presented on its website using identical prompts in the supplementary materials.

%\subsection{Qualitative Evaluation}
%\subsection{Quantitative Comparison}

\subsection{Evaluation on Text-to-3D}
%To comprehensively assess the performance of our decoupled generation method, 
We conducted evaluations in two key aspects: geometry generation and textured model generation. 
%Our evaluation approach encompasses both quantitative metrics and qualitative analysis, complemented by a user study to measure user preferences and perceptions.

\paragraph{Evaluation on Geometry Generation}
Evaluating the quality of generated geometry is a complex problem due to the lack of standard metrics. To objectively evaluate the geometry's quality, we employed a rendering-based approach. 
Specifically, to isolate the geometric attributes from the influence of texture, we rendered the generated geometry with a uniform albedo and then calculated the CLIP score \cite{radford2021learning} using the provided text prompts and CLIP model (vit-g-14). 
This process involved generating 16 different views for each object (a total of $113$ objects) and computing the average score after removing the highest and lowest scores.
%selecting the one with the highest CLIP scores as the representative result. 

\Tref{tab:res_quant} (the first row) shows that two variants of our method achieve the top two values in the average CLIP scores in uniform rendering, demonstrating that our method outperforms existing methods in geometry generation.  Visual results in \fref{fig:res_visual} clearly show that our method can produce 3D content with exceptionally detailed geometry aligned with the text prompts, indicating the effectiveness of the proposed Normal-Depth diffusion model.
%The detailed results of this evaluation are presented in \Tref{tab:res_quant}, demonstrating that our method consistently achieves the highest CLIP scores in uniform rendering. 
%Notably, our approach outperforms existing methods by producing 3D content with exceptionally detailed geometry, all thanks to the innovative normal-depth diffusion models we've introduced.

\paragraph{Evaluation on Textured Model Generation}
In parallel with the geometry evaluation, we assessed the quality of the generated textured models. 
To accomplish this, we computed CLIP scores for the rendered images of the textured models. 
As in the geometry evaluation, we rendered 16 distinct views and computed the average scores. 
\Tref{tab:res_quant} (the second row) shows that the two variants of our method achieve the second and third highest scores, outperforming most of the existing methods.
Our result is slightly lower than that of the ProlificDreamer with a comparison of $31.7099$ vs. $31.8022$.
The reason might be that the ProlificDreamer additionally fine-tunes the diffusion model with LoRA \cite{hu2021lora} during optimization, which might lead to a rendered image that better fits the text prompts.
However, the visual comparison of textured model generation in \fref{fig:res_visual} shows that our method generates much more accurate and detailed models.
These results verify the design of our decoupled text-to-3D generation approach.

%A visual comparison of our text-to-3D generation results is provided in \fref{fig:res_visual}.
%Our method generates visually pleasing textured models with 

\begin{figure}[t] \begin{center}
    \includegraphics[height=0.3\textwidth,width=0.48\textwidth]{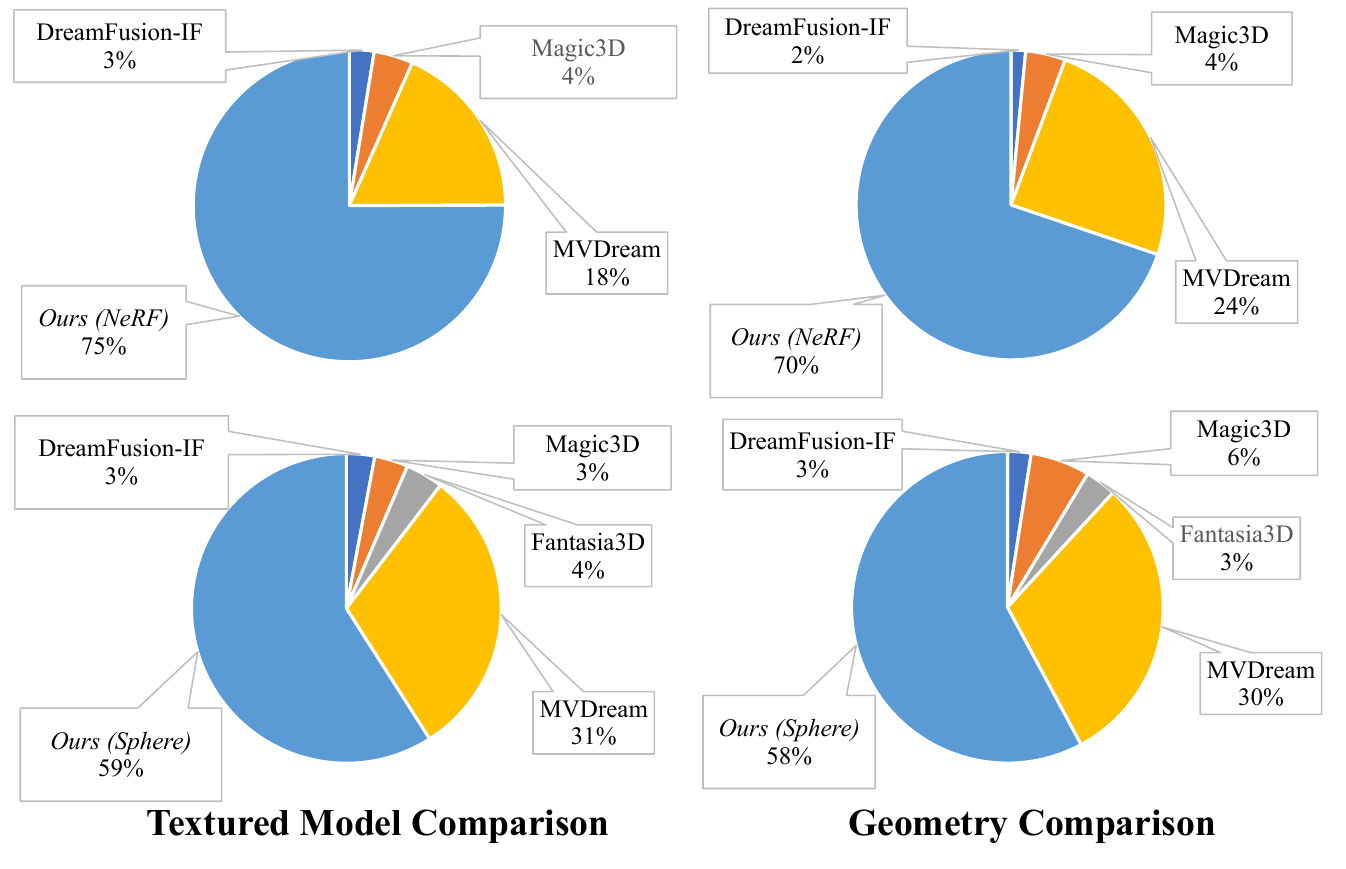}
    \vspace{-7mm}
    \caption{User study for text-to-3D.} \label{fig:userstudy}
    \vspace{-8mm}
\end{center} \end{figure}

\paragraph{User Study}
To further assess the visual quality of the generated 3D models, we conducted a comprehensive user study.
We separately compare the two variants of our method with existing methods.
We collected a set of $87$ prompts from DreamFusions, Sweetdreamer, and MVDream.
119 and 192 participants were involved for the comparison of ``\emph{Ours~(NeRF)} vs. existing methods'' and ``\emph{Ours~(Sphere)} vs. existing methods'' respectively, with each participant undertaking $40$ and $47$ testing.
%For the comparison of Ours (NeRF) with existing methods, 
%For the comparison of Ours (Sphere) with existing methods, 192  participants were involved with each undertake 48 testing.
%involving 100 participants. 
%Each participant was presented with 40 different test cases. 

%Prior to the test, participants were given clear instructions regarding the nature of 3D models, the criteria that define a good 3D model, and the concept of alignment between text descriptions and 3D models.
In each test case, the text prompt, textured models, and normal maps generated by various methods were simultaneously displayed. 
Participants were then tasked to give two votes, one for the best textured model and the other for the best geometry model.
%Rwith selecting the 3D model that they believed exhibited the highest quality and the best alignment with the provided text prompt.
\Fref{fig:userstudy} presents the results of our user study. 
Our method with NeRF representations received $75$\% and $70$ of votes for ``the best textured model''  and ``the best geometry''.
Our method with Sphere initialization received more than $59$\% and $58$ of votes for the two comparisons.
These results show that our method clearly outperforms existing methods on geometry generation and textured model generation, demonstrating the effectiveness of our method.
%When comparing our method, the NeRF-based variant, with other NeRF-based methods, it garnered over 90 percent of the total votes. Similarly, when compared with DMTet-based variants, our method received more than 85 percent of the votes. These outcomes underscore the effectiveness of our approach in generating 3D content characterized by both exceptional visual quality and alignment with textual prompts.
%For a more detailed description of the user study procedure, please refer to the supplementary materials.

\begin{figure} \centering
    \captionof{table}{Ablation study for geometry generation.}
    \resizebox{0.48\textwidth}{!}{
    \begin{tabular}{*{4}{c}}
        \toprule
         & ND Only & ND (w/o LAION) + SD & ND + SD \\
        \midrule
        Geometry CLIP Score &  24.1070 & 24.2601 & 25.8820 \\
        Appearance CLIP Score & 29.5379 & 29.7522 & 31.7099\\
        \bottomrule
    \end{tabular}
    }
    \label{tab:res_quant_ablation}
%\end{table}
%\begin{figure}[tb] \centering

    \includegraphics[width=0.32\textwidth]{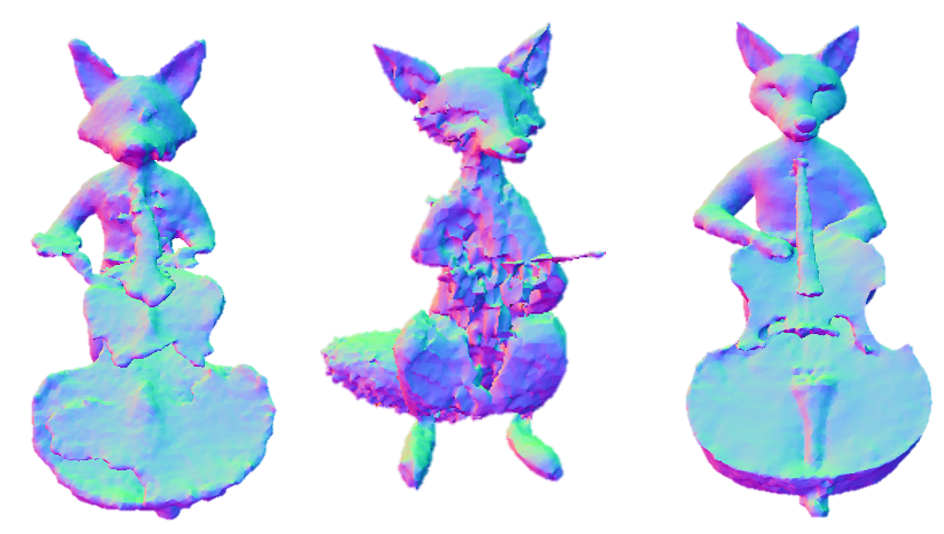}
    \\
    \makebox[0.14\textwidth]{\footnotesize ND only}
    \makebox[0.14\textwidth]{\footnotesize ND (w/o LAION) + SD}
    \makebox[0.14\textwidth]{\footnotesize ND + SD}
    
    % \vspace{1mm}

    \captionof{figure}{Ablation for the Normal-Depth (ND) diffusion model for geometry generation. Prompt: ``\emph{A fox plays a cello}".} 
    \label{fig:res_ablation}
    \vspace{-7mm}
\end{figure}

\subsection{Method Analysis}

\paragraph{Effects of the Normal-Depth Diffusion Model}
To explore the impact of the proposed Normal-Depth diffusion model in the context of text-to-3D, we show results of geometry generation without using the SDS loss from the SD model. 
\Fref{fig:res_ablation} and \Tref{tab:res_quant_ablation} show that only using our Normal-Depth model can robustly generate geometry with a coherent structure. 
When the SD model is incorporated alongside our Normal-Depth model, the resulting geometry exhibits finer details and an improved shape. These findings suggest that our Normal-Depth model serves as a valuable 3D geometric prior for the overall structure, while the SD model excels in generating surface details.
%we compared the generated geometry results using three different approaches: the SD model, our Normal-Depth model (ND), and the combination of our normal-depth model with the SD model (ND+SD).

%It is observed that optimizing geometry with the SD model often fails when initialized with a sphere (see \fref{fig:res_ablation}).
%Conversely, our normal-depth model consistently generates geometry with a robust and coherent structure. 

\paragraph{Effects of Pre-training on LAION dataset}
To evaluate the impact of pre-training on the LAION-2B dataset using normal and depth generated by existing methods  \cite{bae2021estimating,ranftl2020towards}, we conducted a comparison with a baseline model that directly fine-tunes on the synthetic Objaverse dataset \cite{deitke2023objaverse}. The resulted baseline text-to-3D model is denoted as \hbox{\emph{ND~(w/o~LAION)+ SD}}. 
Figure \ref{fig:res_ablation} illustrates that when the Normal-Depth model is fine-tuned solely on the synthetic dataset, its generalization ability significantly deteriorates. It struggles to generate content that aligns with the user prompts, and the quality of the generated geometry is notably inferior. In contrast, our method, which involves pre-training on the expansive LAION dataset, successfully preserves its generalization ability and produces superior results, which is also evidenced in \Tref{tab:res_quant_ablation}.

\begin{figure}[tb] \centering
    \includegraphics[width=0.48\textwidth]{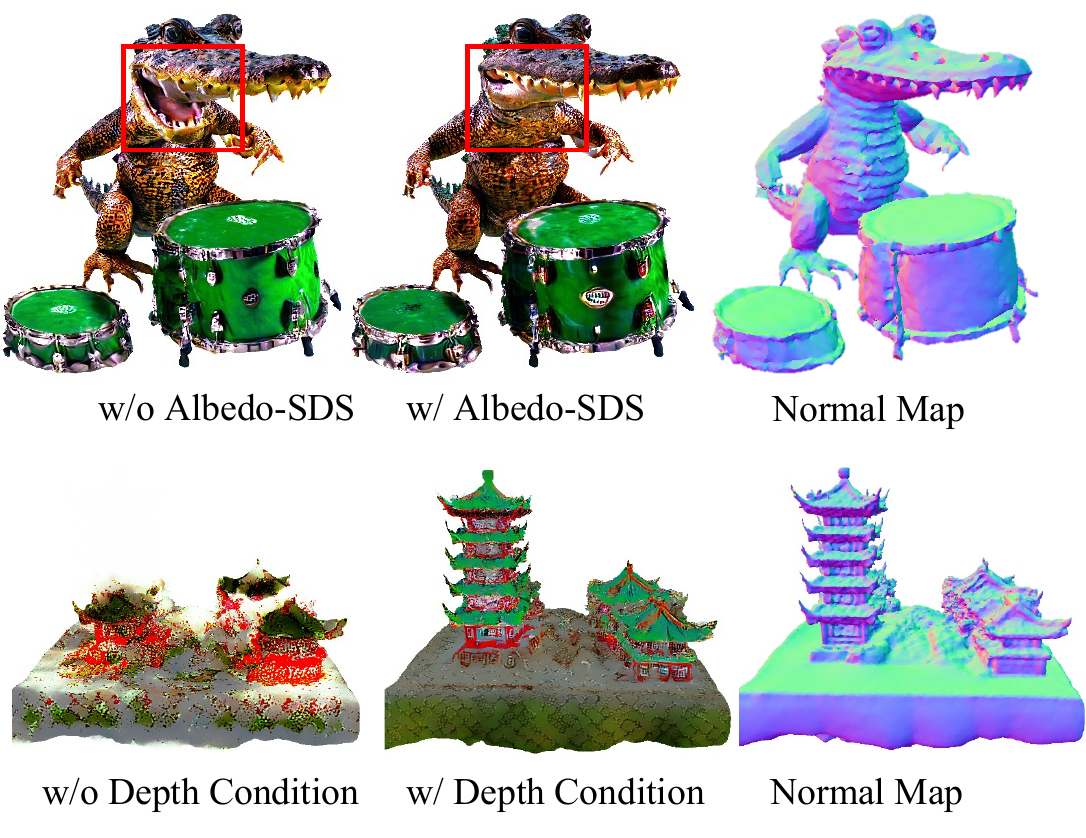}
    % \vspace{-0.2em}
    \vspace{-7mm}
    \caption{Ablation results for the albedo diffusion model.} 
    \label{fig:res_ablation_albedo}
\end{figure}

\begin{figure}[tb] \centering
    \vspace{-3mm}
    %\makebox[0.07\textwidth]{\scriptsize w/o Albedo$_{SDS}$}
    %\makebox[0.07\textwidth]{\scriptsize Shading}
    %\makebox[0.07\textwidth]{\scriptsize w Albedo-SDS}
    %\makebox[0.07\textwidth]{\scriptsize w/o Albedo-SDS}
    %\makebox[0.07\textwidth]{\scriptsize Shading}
    %\makebox[0.07\textwidth]{\scriptsize w Albedo-SDS}
    %\\
    \includegraphics[width=0.235\textwidth]{ 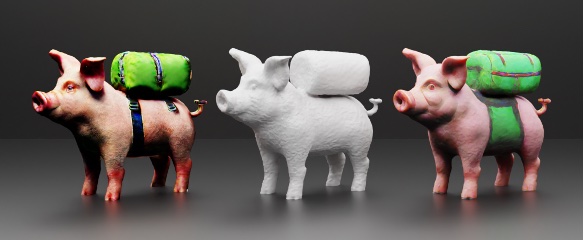}
    \includegraphics[width=0.235\textwidth]{ 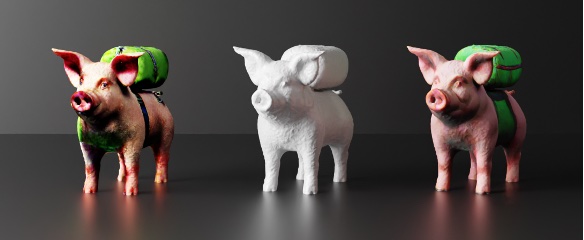}
    \\
    \caption{Relighting results. From left to right: results of model w/o albedo diffusion, shading, and model w/ albedo diffusion.} \label{fig:res_brdf}
    \vspace{-6.5mm}
\end{figure}

\paragraph{Effects of Albedo Diffusion Model}
\Fref{fig:res_ablation_albedo} shows that the albedo diffusion model can effectively improve the generated texture and lead to a more accurate appearance.
Without the depth condition, the generated texture fails to align with the underlying geometry, highlighting the importance of depth condition in albedo diffusion model.
With the inclusion of the albedo diffusion model, the generated albedo exhibits reduced shadows and specular highlights, leading to a more realistic relighting results (see~\fref{fig:res_brdf}). 
% of the proposed model in achieving a coherent and visually appealing result.
%The relighting results under novel illumination in \fref{fig:res_brdf} demonstrate that the generated materials is more suitable for relighting.
%
%\paragraph{Comparison of DMTet and NeRF Representation}
%\begin{figure}[t] \begin{center}
%    \includegraphics[height=0.2\textwidth,width=0.48\textwidth]{example-image}
%    \caption{Visualization of material decomposition.} \label{fig:res_brdf}
%\end{center} \end{figure}

\section{Conclusion}
\label{sec:Conclusion}
In this work, we presented a generalizable approach to 3D generation through a Normal-Depth diffusion model, trained extensively on real-world data before undergoing fine-tuning with synthetic datasets. 
We also introduced a depth-conditioned albedo diffusion model that facilitates the separation of material attributes and lighting effects.
Our models seamlessly integrate into current text-to-3D pipelines and demonstrate compatibility with the NeRF and DMTet representations.
Extensive experiments show that our method achieves state-of-the-art text-to-3D results in both geometry and appearance modeling.

\paragraph{Future Work}
Our current method predominantly focuses on object-level 3D generation. Moving forward, we aim to extend our Normal-Depth diffusion model to address text-to-scene generation challenges. Additionally, an interesting direction for future research is the development of an appearance prior model that regularizes the specular component of 3D content.

\paragraph{Acknowledgement} We would like to express our special gratitude to Rui Chen for the valuable discussion in training Fantasia3D and PBR modeling. Additionally, we extend our heartfelt thanks to Chao Xu for his assistance in conducting relighting experiments.

%Our current method mainly focuses on the object-level 3D generation. In the future, we would like to extend our normal-depth diffusion model to tackle the problem of text-to-scene generation.
%Additionally, developing a appearance prior model to regularize the specular component of 3D content is an interesting direction for future research.
%
%In this work, we attempt to enhance appearance modeling by applying data-driven regularization to the generated albedo, utilizing an albedo diffusion model. Future directions include developing a prior model to regularize the specular component of 3D content. Additionally, extending our proposed Normal-Depth diffusion model to enable scene-level generation presents an interesting direction for further research.

%\paragraph{Future Work} In this work, we make an attempt to improve appearance modeling by imposing data-driven regularization on the generated albedo based on an albedo diffusion model. It would be interesting to develop a prior model to regularize the specular component of the 3D contents. 
%Moreover, extending the proposed Normal-Depth diffusion model to scene-level generation is another interesting future work.
%Moreover, Our current method mainly focuses on the object-level 3D generation. In the future, we would like to tackle the more challenging problem of text-to-scene generation by leveraging the capabilities of our proposed normal-depth diffusion model.

\clearpage
%%%%%%%%% REFERENCES
{\small

\bibliographystyle{ieee_fullname}
}

\clearpage
\appendix

% \section{Supplementary Video}
% Please kindly check the attached video for $360^\circ$ rendering results of the generated 3D content.

\section{More Details for Normal-Depth Diffusion}

\subsection{Pre-training on the LAION Dataset}

\begin{figure*}[tb] \centering
   \includegraphics[width=\textwidth]{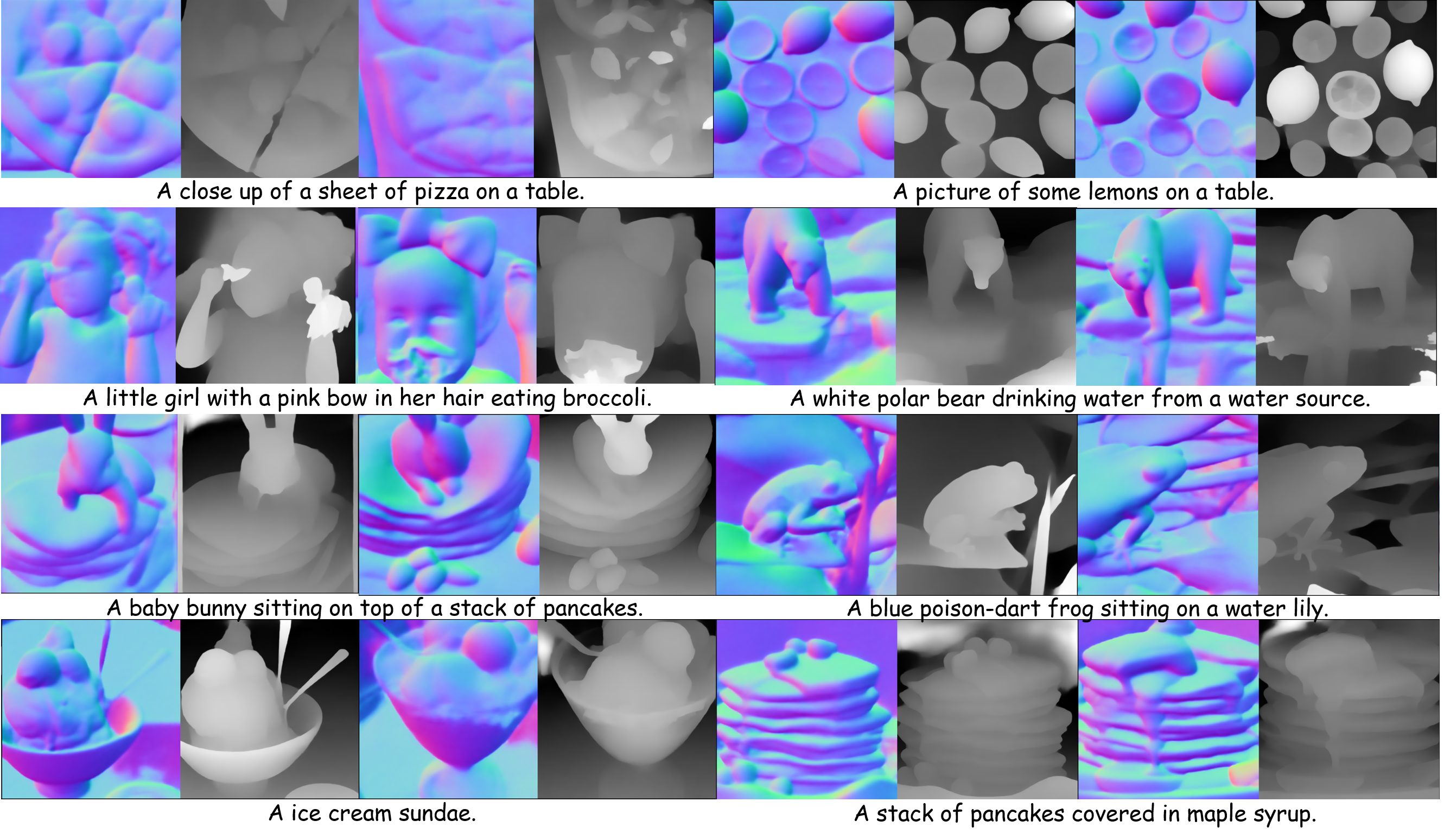}
    % \vspace{-3mm}
    \caption{Text to Normal-Depth sampling results.} 
    \label{fig:txt2ndsampling}
\end{figure*}

\paragraph{Training of VAE}

We initialize the parameters of our model with the pre-trained weights of SD 2.1. Specifically, we modify the input channels of the VAE from 3 to 4, and the weights of the newly added channel are initialized as the average of the original weights. We do not modify the number of channels in the latent space of our VAE, which remains at 4 channels.

For VAE fine-tuning, we randomly sample the training data from LAION-Aesthetics V1~\cite{schuhmann2022laion}, selecting data with aesthetics scores larger than 8.0, considering the need for high-quality training data. The training dataset consists of 8 million training samples.

We apply center cropping on the training images and resize them to $384\times384$. We employ random rotation and flipping of the images to augment our training dataset.
Subsequently, we use augmented images as inputs to the monocular prior models~(NormalBae~\cite{bae2021estimating} and Midas-3.1~\cite{ranftl2020towards}) to obtain corresponding estimates of the normal and depth. 
Notably, we perform depth normalization on the predicted depth values, scaling them from -1 to 1. This normalization step is necessary since the original depth predictions are in the form of real values. These estimated images are then resized to a resolution of $256\times256$ and serve as the input for the VAE.

During the training phase, following the latent diffusion model~\cite{rombach2022high}, we employ the Adam optimizer with a learning rate of 5e-5 to optimize our VAE model. To ensure a well-behaved latent space, we incorporate KL regularization loss. Moreover, to further improve the quality of the generated images, we train an auxiliary discriminator on the output of the VAE. The weights of the KL regularization and discriminator are set to 1e-6 and 0.5, respectively. For each iteration, the batch size is set to 1024. The training process is conducted on 8 A100-80G GPUs for two weeks, reaching a total of 100K iterations.

\paragraph{Training of UNet Diffuser} Notably, we maintain the original structure of the UNet model as the channels of the latent remain unchanged. During the training phase, we randomly sample data from the Laion-2B-en dataset. We perform a center crop on the sampled images and resize them to a resolution of 512 pixels. Subsequently, these images are passed through the monocular prior models and the encoder of the trained VAE. The output of this process serves as the input of the UNet model. 

We follow the strategies utilized in the latent-diffusion model~\cite{rombach2022high} to train our Normal-Depth diffusion model.
Specifically, we first train our Normal-Depth diffusion model using the entire Laion-2B-en dataset. 
After 121,000 iterations of training, we proceed to sample our data from the ``Laion-Aesthetics v2 5+'' subset. This subset contains images with aesthetics scores greater than 5, and we only select images with an unsafety probability higher than 0.1.
The fine-tuning process continues for about 167,000 iterations at a resolution of $512 \times 512$. 
To enable classifier-free guidance sampling, we incorporate a 10\% dropping of the text-conditioning. We utilize the Adam optimizer with a learning rate of 1e-4 to optimize our Normal-Depth diffusion model. For each iteration, the batch size is set to 1024. 
This process costs 11,520 GPU hours.
%This process was conducted on 32 A100-80G GPUs for about 20 days. 

 \paragraph{Text to Normal-Depth Sampling} \Fref{fig:txt2ndsampling} shows the sampling results of our Normal-Depth diffusion model trained on the Laion-2B datasets. As shown in the figure, the sampled normal and depth are not only highly consistent with the textual description but also with high quality.
 We set the classifier-free guidance scale to 7.5, with 50 DDIM~\cite{song2020denoising} steps.

\subsection{Fine-tuning on Synthetic Dataset}%

\paragraph{Multi-view Normal-Depth Diffusion Fine-tune}

Thanks to the open-source large-scale Objaverse dataset~\cite{deitke2023objaverse}, we fine-tune our Normal-Depth diffusion model on Objaverse to improve its performance for 3D generation tasks. 
To avoid the Janus problem, following MVDream~\cite{shi2023mvdream}, we finetune the Normal-Depth Diffusion model pretrained using multi-view diffusion. Particularly, we use 4 images of orthogonal camera views as the input for the diffusion model and apply a two-layer MLP to embed the extrinsic camera matrix, which is added as a residual to the time embedding.
%To avoid the Janus problem, following MVDream~\cite{shi2023mvdream}, we finetune the Multi-view Normal-Depth Diffusion model pretrained on our Normal-Depth Diffusion model. Particularly, we use 4 orthogonal camera views' images as diffusion input and apply a two-layer MLP to embed the extrinsic camera matrix, which is added as a residual to the time embedding.

\begin{figure}[tb] 
\centering
\includegraphics[width=0.48\textwidth]{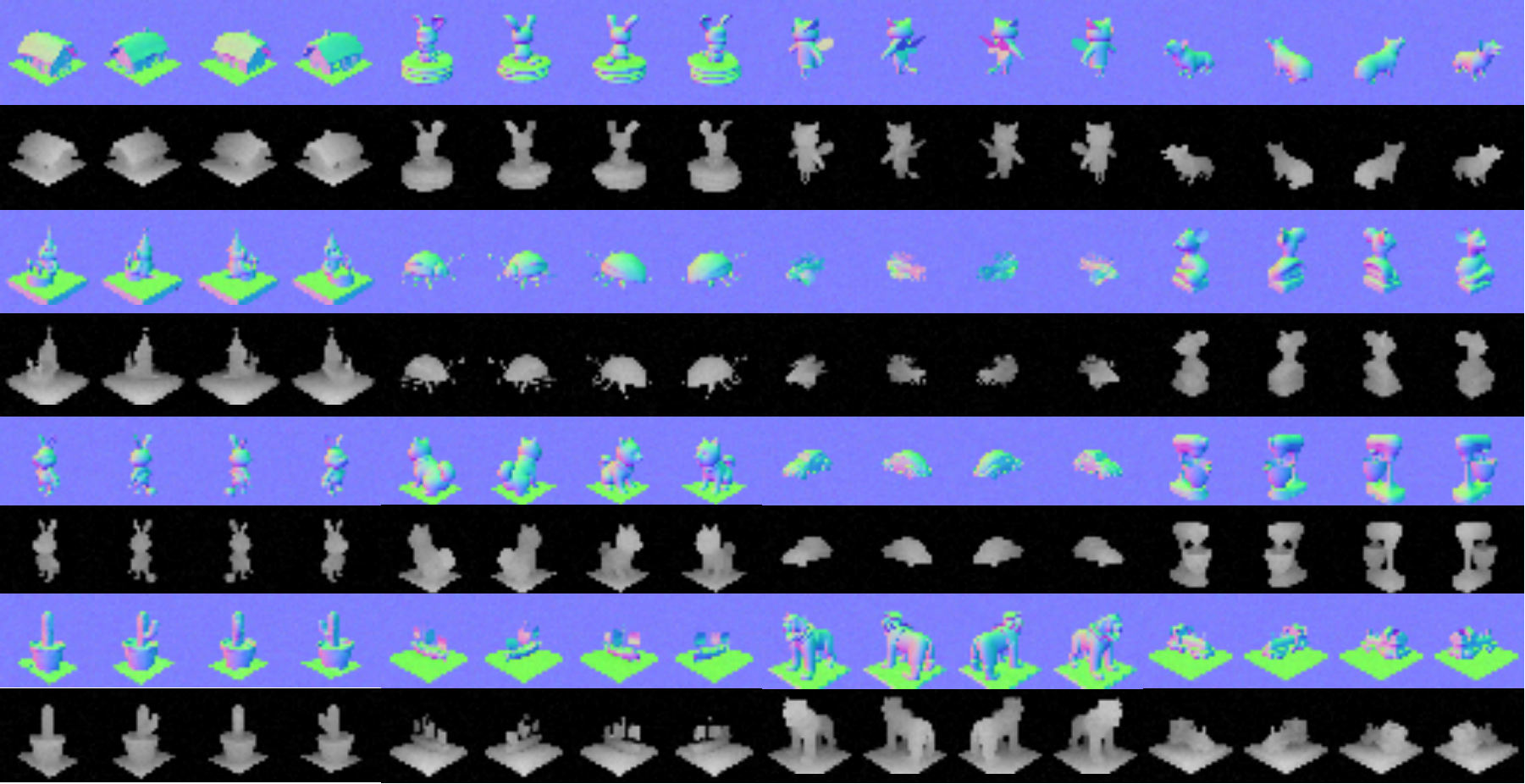}
\caption{Sampling results of our Normal-Depth diffusion model fine-tuned on the Objaverse dataset.} 
\label{fig: mv-nd sampling}
\end{figure}

\Fref{fig: mv-nd sampling} illustrates the sampling results of our multi-view Normal-Depth Diffusion model, where the classifier-free guidance scale is set to 10 and the negative prompt is set to null.

\paragraph{Depth Normalization}

For a synthetic dataset, we can directly obtain its depth values. Since our Normal-Depth Diffusion process involves normalized depth during training, it is necessary to normalize the range of synthetic depth. To better normalize the depth values, we can introduce a near plane and a far plane to restrict the depth range to [-1, 1]. By defining these planes, we can map the actual depth values to the normalized range. Considering our synthetic data is confined within a 0.5 uint cubic volume, we define the distance from the object's center point to the near plane and far plane as $0.5\sqrt{3}$. After defining the near and far planes, we can normalize the depth values of our synthetic data.

However, the process of normalizing depth is not trivial. Midas~\cite{peng2017depth} estimates depth in the form of disparity, which is a relative measure of the difference in pixel coordinates and is inversely related to depth.
% between the left and right views of a stereo image pair

\begin{figure}[tb] 
\centering
\includegraphics[width=0.4\textwidth]{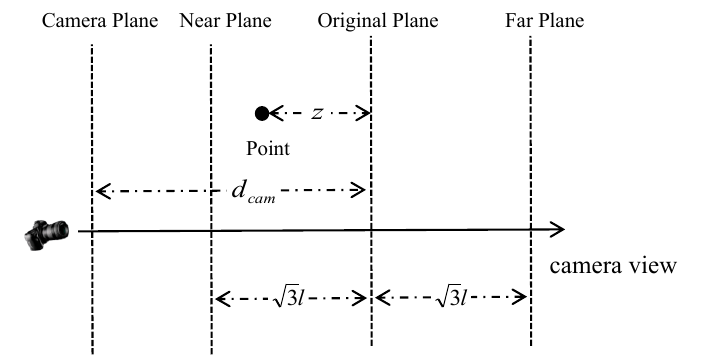}
\caption{Depth normalization figure.} 
\label{fig:depth-normalization-fig}
\end{figure}

One straightforward approach is to normalize the disparity of synthetic data directly.  For simplicity, we assume that the synthetic object is located at the coordinate origin. The equation for normalizing the disparity is as follows:
\begin{equation}
\begin{aligned}
    \textrm{Disp}(z) &=\frac{\frac{1}{d_{\text{cam}}-z}-\frac{1}{d_{\text{cam}}+\sqrt{3}l}}{\frac{1}{d_{\text{cam}}-\sqrt{3}l}-\frac{1}{d_{\text{cam}}+\sqrt{3}l}} \\
    &=\frac{(\sqrt{3}l+z)\cdot(d_{\text{cam}}-\sqrt{3}l)}{2\sqrt{3}l\cdot(d_{\text{cam}}-z)}.
\end{aligned}
\end{equation}
where $z$ represents the depth value between the given point and the original plane, ${d_{\text{cam}}}$ denotes the depth value from the origin plane to the camera, and $l$ represents the size of the cube that confines the object, as illustrated in \fref{fig:depth-normalization-fig}.

From the equation, it is evident that the normalized disparity lacks the desirable property of scale invariance. Consider an example: for a fixed camera distance ($d_{\text{cam}}$), if we double the values of $l$ and $z$, the resulting values will differ from the original $l$ and $z$ values. Similarly, for the same values of $l$ and $z$, different camera distances will yield different normalized disparities.

During the optimization process of text to 3D, we randomly sample different camera distances. This randomness further exacerbates the lack of scale invariance, introducing significant noise into the optimization process. As a consequence, achieving accurate results becomes more challenging due to the varying scales, which can adversely affect the convergence and stability of the optimization algorithm.

To address the aforementioned issue, we propose the use of \textit{reverse depth} as an alternative to disparity normalization. The reverse depth is defined as follows:

\begin{equation}
\begin{aligned}
    \textrm{RevDepth}(z) &=\frac{(d_{\text{cam}}+\sqrt{3}l)-(d_{\text{cam}}-z)}{(d_{\text{cam}}+\sqrt{3}l)-(d_{\text{cam}}-\sqrt{3}l)} \\
    &=\frac{\sqrt{3}l+z}{2\sqrt{3}l}.
\end{aligned}
\end{equation}
From the equation, it is obvious that the normalization value is independent of $d_{\text{cam}}$ and remains unchanged when the variables $l$ and $z$ scale proportionally. For the sake of simplicity, in the following content, the depth refers to normalized reverse depth.

% Figure~\ref{fig:camparison depth-normalization} demonstrates the comparison results between using disparity and reverse-depth normalization.
% For the sake of simplicity, in the following content, the depth refers to normalized reverse depth.
\begin{figure}[tb] 
\centering
\includegraphics[width=0.48\textwidth]{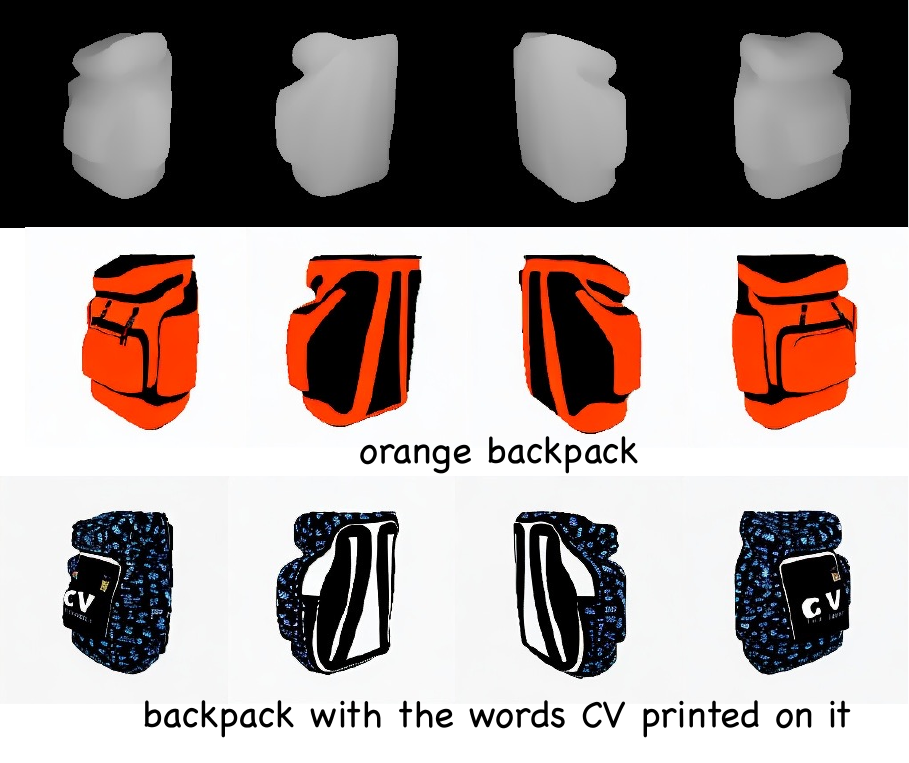}
\caption{Depth-Condition Albedo diffusion Model.}
\label{fig:mv-albedo-sampling}
\end{figure}

\section{More Details for Albedo Diffusion Model}
\paragraph{Depth-Conditioned Albedo Diffusion Fine-tuning}

Regarding the training of the albedo model under the depth condition, we resize the normalized depth image and concatenate it with the latent space of the VAE. It means the number of channels in the UNet's input expands from 4 to 5. We initialize the parameters of the Albedo-Diffusion Model with the pre-trained weights of SD 2.1. For the additional dimension in the input channel of the UNet, we set its weights to zero values. 
To alleviate the multi-face problem in the generated texture, we also employ the same multi-view strategy used in the multi-view Normal-Depth Diffusion model to train our albedo diffusion model (see~\fref{fig:mv-albedo-sampling} for sampling results).

% \begin{figure*}[h] \centering
%     \includegraphics[width=\textwidth,height=0.4\textwidth]{example-image}
%     \caption{Image 1.} \label{fig:}
% \end{figure*}

% \begin{table}[h]\centering
%     \caption{A simple table with a header row.}
%     \label{tab:table1}
%     \resizebox{0.48\textwidth}{!}{
%     \large
%     \begin{tabular}{*{10}{c}}
%         \toprule
%        Data & Size &  2-Exp & 3-Exp &  4-Exp & 5-Exp &  6-Exp &  7-Exp \\
%         \midrule
%         A & $1280\times 720$ & 1 & 2 & 3 & 4 & 5 & 4 \\
%         B & $1280\times 720$ & 1 & 2 & 3 & 4 & 5 & 4 \\
%         Ours & $4096\times 2168$ & 2 & 3 & 4 & 6 & 5 & 4 \\
%         \bottomrule
%     \end{tabular}
%     }
% \end{table}

% origin
\section{More Details for Geometry Generation}
Our Normal-Depth diffusion model can be applied to optimize DMTet and NeRF. 
To better verify the effectiveness of our Normal-Depth Diffusion model, we design two model variants for text-to-3D. 
The first model, denoted as \emph{Ours (Sphere)}, initializes the DMTet with a Sphere for geometry generation. The second model, denoted as \emph{Ours (NeRF)}, first optimizes a NeRF with our Normal-Depth diffusion model and then converts the NeRF to DMTet as an initialization for geometry generation.

In the paper, the geometry generation loss function is defined as:
\begin{align}
    \label{eq:sds_geo}
    \mathcal{L}_\text{Geo} = \lambda_\text{SD} \mathcal{L}_{\text{SDS}-\text{Normal}}^\text{SD} + \lambda_\text{ND} \mathcal{L}_{\text{SDS}-\text{ND}}^\text{ND},
\end{align}

\paragraph{More Details for \emph{Ours~(Sphere)}}
%For the sake of clarity, we refer to the variant obtained by integrating our Normal-Depth Diffusion Model into Fantasia3D~\cite{chen2023fantasia3d} as the Ours~(Sphere) pipeline. 
%All that is required is an additional parallel branch to incorporate Normal-Depth priors for supervising the geometry modeling in the original pipeline.

The geometry optimization in \emph{Ours~(Sphere)} consists of two stages: a coarse shape optimization stage and a refinement stage. For the coarse shape optimization stage, we follow the approach of Fantasia3D, where we directly resize the rendered normal and depth maps as the latent space features to quickly optimize to obtain a coarse shape. In the refinement stage, we utilize the latent features obtained from the VAE to enhance the details of geometry.

Specifically, our DMTet is initialized from a sphere. In terms of coarse shape optimization, we set $\lambda_\text{SD}$ and $\lambda_\text{ND}$ to 0.5 and 1.0, respectively. When it comes to refinement optimization, both $\lambda_\text{ND}$ and $\lambda_\text{SD}$ are set to 1.0. The negative prompt ``low quality'' is used in both diffusion models. The classifier-guided scale is set to 100 and 50 in SD 2.1 and the Normal-Depth diffusion model, respectively. For the time sampling schedule, we adopt a uniform sampling strategy of annealing from [0.5, 0.98] to [0.05, 0.5] when we switch from the coarse to refinement stage.

\emph{Ours~(Sphere)} is optimized on a single Nvidia A100-80G GPU for 3000 iterations (1500 iterations for the coarse stage and 1500 iterations for the fine stage), where we use the AdamW optimizer with learning rates of 1e-3. The batch size is set to 8, and the entire geometry optimization process takes about 40 minutes.

\paragraph{More Details for \emph{Ours~(NeRF)}}
The geometry optimization in \emph{Ours~(NeRF)} also consists of two stages. 
In the coarse shape optimization stage, we adopt the rendered normal and depth maps as latent space input of SD 2.1 and the Normal-Depth diffusion model without VAE encoding to quickly optimize to obtain the coarse shape. 
% we found that utilizing rendered color image and opacity map as latent space input of SD V2.1 could get better results for coarse shape generation.
In the fine detail refinement stage, we adopt encoding features obtained from VAE as latent space input of SD 2.1 to get a detailed shape.

\begin{figure}[t]
    \centering
    \includegraphics[width=0.48\textwidth]{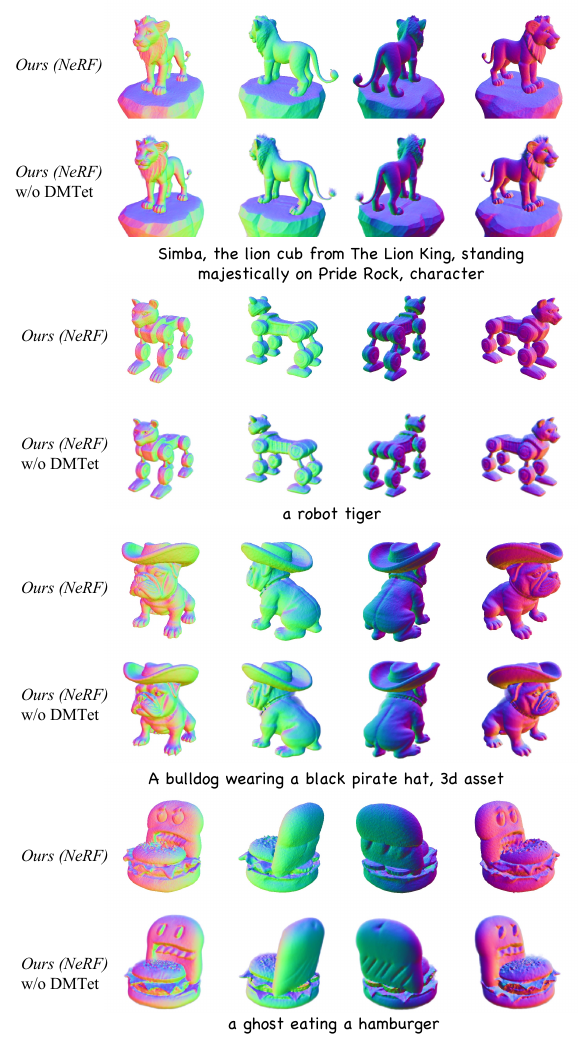}
    \caption{The visual geometric comparison results about with and without DMTet refinement stage in \emph{Ours~(NeRF)}.}
    \label{fig:nerf_compare}
\end{figure}

\emph{Ours~(NeRF)} is optimized on a single Nvidia A100-80G GPU for 5000 iterations (1500 iterations for the coarse stage and 3500 iterations for the fine stage), where we use the AdamW optimizer with learning rates of 1e-3 except for the hash encoding module using 1e-2.
For the time sampling schedule, we adopt a uniform sampling strategy of annealing from [0.5, 0.98] to [0.05, 0.5] at the 3000th iteration. We set $\lambda_\text{SD}$ to 1.0 and $\lambda_\text{ND}$ annealing 10 to 2 at the 3500th iteration.
The classifier-guided scale is set to 50 in both SD 2.1 and the Normal-Depth diffusion model.
We utilize a multi-resolution strategy to train NeRF efficiently, the rendering resolution increases from 64 $\times$ 64 to 256 $\times$ 256 at the 3000th iteration, and the batch size decreases from 8 to 4. The entire geometry optimization takes about 40 minutes.

To reduce geometric artifacts when converting NeRF to DMTet representations, we optimize the DMTet with an additional 3000 iterations to refine the geometry. This is done using the same strategy adopted in the fine detail stage, except the rendering resolution is increased to 512. The refinement optimization takes about 20 minutes.

As illustrated in Figure~\ref{fig:nerf_compare}, the optimized geometry of NeRF using our Normal-Depth diffusion model is already of very high quality, clearly demonstrating the effectiveness of our Normal-Depth diffusion model in optimizing both the DMTet and NeRF representations. After undergoing DMTet conversion and subsequent refinement, the surface details are further enhanced. However, for geometry types that are more suitably represented by a density volume (e.g., hair and smoke), the geometry shape tends to be better without the DMTet conversion.

%for generating certain objects that encompass hair, smoke, and other elements more suitably represented by NeRF, the geometry shape would be better without DMTet conversion.
%finer details in the shapes after
%undergoing DMTet conversion and subsequent refinement. 

\paragraph{Camera Sampling} During the training process, we randomly sample the elevation angle between 5 degrees and 30 degrees and uniformly sample the camera distance between 1.5 and 1.9. In addition, for the sampling of azimuth angles, we follow the sampling approach from MVDream~\cite{shi2023mvdream}, where we consecutively sample four orthogonal viewpoints. 

%For \emph{Ours~(Sphere)} optimization, we set the side length of our volume cube to be 1 and initialize the radius of the sphere to be 0.5.

\section{More Details for Appearance Modeling}
The classifier guidance scale is set to 10 for the depth condition Albedo-Diffusion model, while its value is 100 for the original SD 2.1. We harness the same camera sampling strategy in geometry appearance modeling. In terms of the time sampling schedule, we adopt a uniform sampling strategy from [0.02, 0.98].

Our appearance model is optimized on a single Nvidia A100-80G GPU for 3000 iterations. The batch size is set to 8, and the AdamW optimizer with learning rates of 1e-2 is utilized to update the model parameters.

\section{More Details for the User Study}
In the user study compared with the baseline, our main focus is to evaluate the quality of the generated geometry and the textured models. Regarding geometry, we primarily compare whether the geometry is complete, if the fine details appear natural, and whether there are significant artifacts. 
For textured models, our comparison is based on the naturalness of the generated textures and the alignment between the textured model and the textual descriptions. At the beginning of each questionnaire, we will provide an illustrative example to explain what is meant by ``visual-textual matching'' and how to evaluate the quality of the generated geometry.

\begin{figure}[t]
    \centering
    \includegraphics[width=0.48\textwidth]{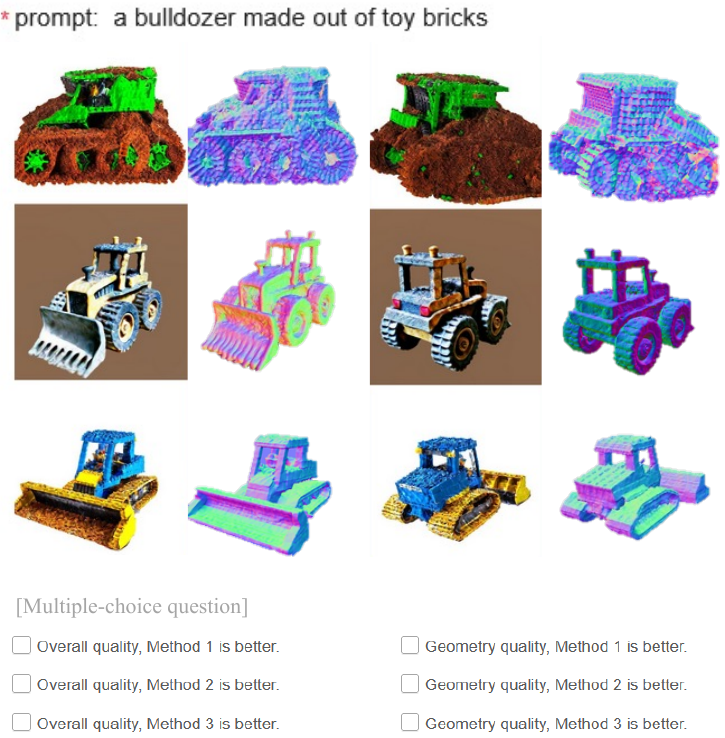}
    \caption{Example of the interactive interface for the user study.}
    \label{fig:user-study-gallery}
\end{figure}

The interface of our user study is shown in \fref{fig:user-study-gallery}, where we display the results of different methods in each row and randomly shuffle the order of the methods to avoid introducing bias. For each row, we display four images, which consist of color and normal maps captured from two camera views that are 180 degrees apart.

We distribute our questionnaire to graduate students and professionals working in the field of 3D, such as model designers and engineers from technology companies. 
All the prompts used in our user study are presented in the attached \emph{txt} file.

\begin{figure}[t]
    \centering
    \includegraphics[width=0.48\textwidth]{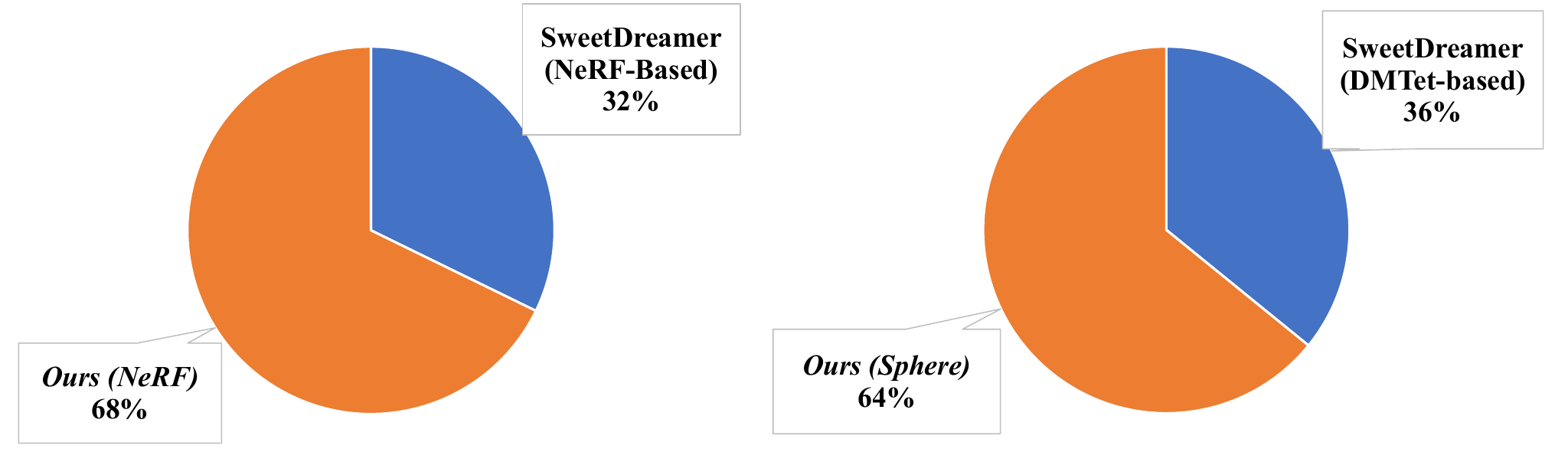}
    \\
    \makebox[0.21\textwidth]{\footnotesize \emph{Ours~(NeRF)} vs. NeRF-Based}
    \makebox[0.21\textwidth]{\footnotesize \emph{Ours~(Sphere)} vs. DMTet-Based}
    \caption{The user study for the comparison with SweetDreamer.}
    \label{fig:sweetdreamer-user-study}
\end{figure}

\begin{figure}[h]
    \centering
    \includegraphics[width=0.48\textwidth]{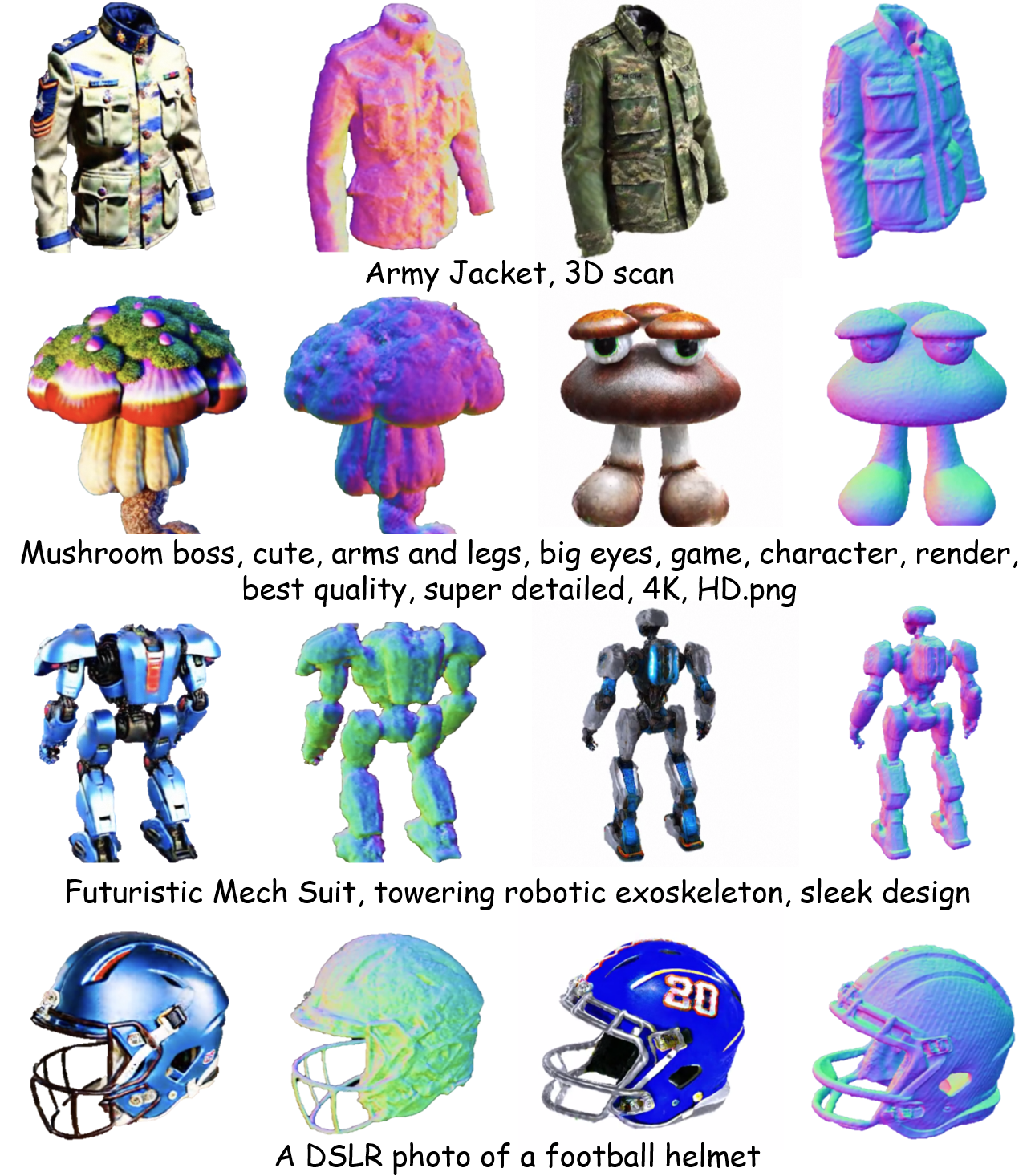}
    \caption{Comparison between SweetDreamer~(NeRF-based) and \emph{Ours~(NeRF)}. In each row, from left to right show the rendered image and normal map of the SweetDreamer, followed by the rendered image and normal map of our method.}
    \vspace{-4mm}
    \label{fig:sweetdreamer-NeRF}
\end{figure}

\begin{figure}[h]
    \centering
    \includegraphics[width=0.48\textwidth]{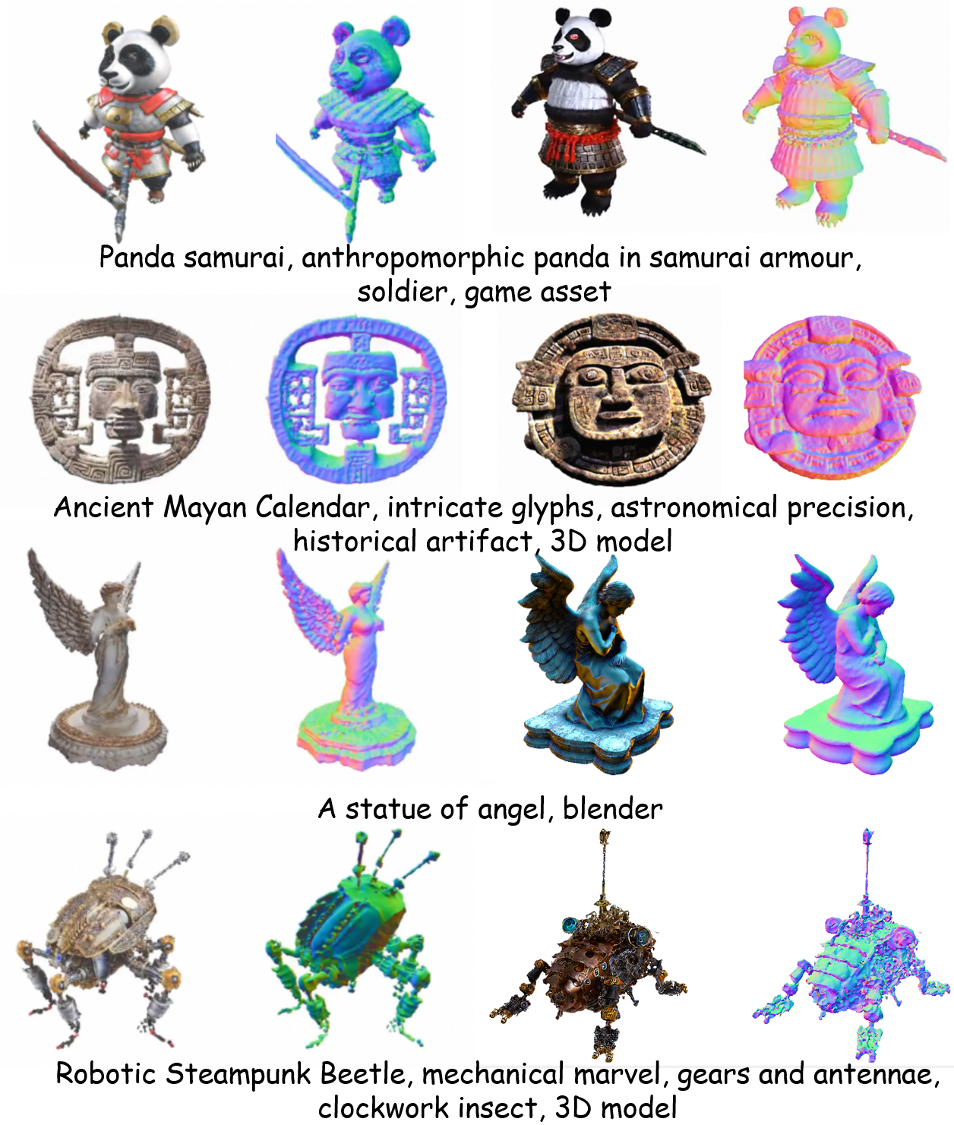}
    \caption{Comparison between SweetDreamer~(DMTet-based) and \emph{Ours~(Sphere)}. In each row, from left to right show the rendered image and normal map of the SweetDreamer, followed by the rendered image and normal map of our method.
    }
    % \vspace{-2mm}
    \label{fig:sweetdreamer-dmtet}
\end{figure}

\section{More Results}
\subsection{Comparison with SweetDreamer}
In comparison to SweetDreamer~\cite{li2023sweetdreamer}, a concurrent work that does not have publicly available code, we selected 40 text prompts from their project page or paper~(20 prompts from their NeRF-based approach and 20 from their DMTet-based approach). Since the DMTet-based method from SweetDreamer visualizes the shape with the \emph{shading normal} instead of the \emph{geometry normal}, it is not feasible to directly compare the geometry quality. Therefore, we focus solely on evaluating the overall quality of the textured models.

\Fref{fig:sweetdreamer-user-study} illustrates the results of the user study comparing our method with SweetDreamer. In our user study, a significant majority of participants expressed a preference for our method.
Specifically, when compared with SweetDreamer's NeRF-based approach, 68\% of users selected \emph{Ours~(NeRF)} as their preferred choice. When compared with SweetDreamer's DMTet-based approach, 64\% of users chose \emph{Ours~(Sphere)} as their preferred option.
This outcome demonstrates that our method outperforms SweetDreamder in 3D generation.

%Specifically, Our~(\emph{NeRF}) outperforms SweetDreamer by 36\% when compared to NeRF-based approach proposed by SweetDreamer. 
%Additionally, when compared to DMTet-Based method, Our~(\emph{Sphere}) shows an improvement of approximately 28\% over SweetDreamer.

\Fref{fig:sweetdreamer-NeRF} and \Fref{fig:sweetdreamer-dmtet} present the comparisons with SweetDreamer~(NeRF-based) and SweetDreamer~(DMTet-based), respectively. We can observe that our method generates better geometry compared to SweetDreamer.

%With respect to qualitative comparison results, both \Fref{fig:sweetdreamer-NeRF} and \Fref{fig:sweetdreamer-dmtet} demostrate the comparison with SweetDreamer~(NeRF-based) and SweetDreamer~(DMTet-based), respectively.  From the images, we can observe that our method generates better geometry compared to SweetDreamer.
\begin{figure}[t] \centering
    \includegraphics[width=0.42\textwidth]{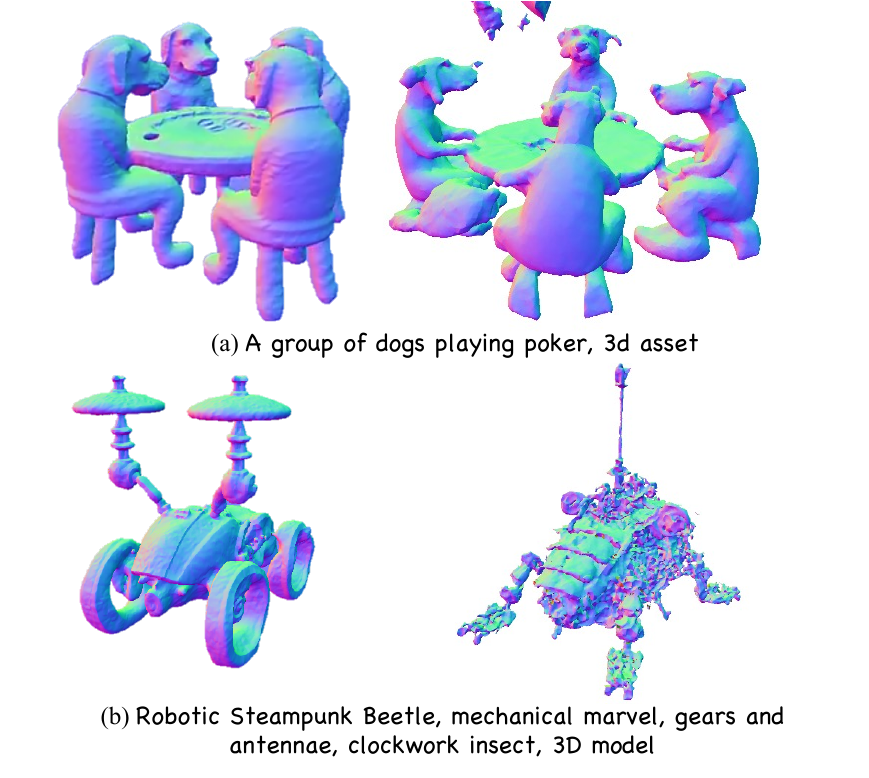}
    \\
   \vspace{-0.6em}
    \makebox[0.21\textwidth]{\footnotesize \emph{Ours~(NeRF)}}
    \makebox[0.21\textwidth]{\footnotesize \emph{Ours~(Sphere)}}
   %\vspace{-2em}
   \caption{Comparison between \emph{Our~(NeRF)} and \emph{Our~(Sphere)}.}    \vspace{-5mm}
   \label{fig:comparison-nerf-with-sphere}
\end{figure}

\subsection{Discussion for
\emph{Ours~(Sphere)} and \emph{Ours~(NeRF)}}
To better understand the behavior of our method, we discuss differences between the  \emph{Ours~(Sphere)} and \emph{Ours~(NeRF)}.
In our experiments, we found that DMTet initialized from NeRF and from a sphere has advantages for different cases.

In terms of NeRF initialization, it is easier to generate scenes with multiple objects, e.g., ``a group of dogs playing poker''. \Fref{fig:comparison-nerf-with-sphere}~(a) demonstrate the comparison results between \emph{Ours~(NeRF)} and \emph{Ours~(Sphere)} in this case. 
However, compared to optimization from a sphere, the geometry from NeRF initialization tends to be smoother, making it difficult to generate surfaces with highly detailed structures, as shown in \Fref{fig:comparison-nerf-with-sphere}~(b).   

%In the future, we would like to investigate better strategy by combining the merits of both initialization methods to  achieved improved results.
In the future, we aim to devise a novel hybrid representation that combines both initialization methods. This approach will allow us to leverage the strengths of each initialization and potentially yield improved results.

\subsection{More Visual Results}
\begin{figure*}[h] \centering
    \includegraphics[width=\textwidth]{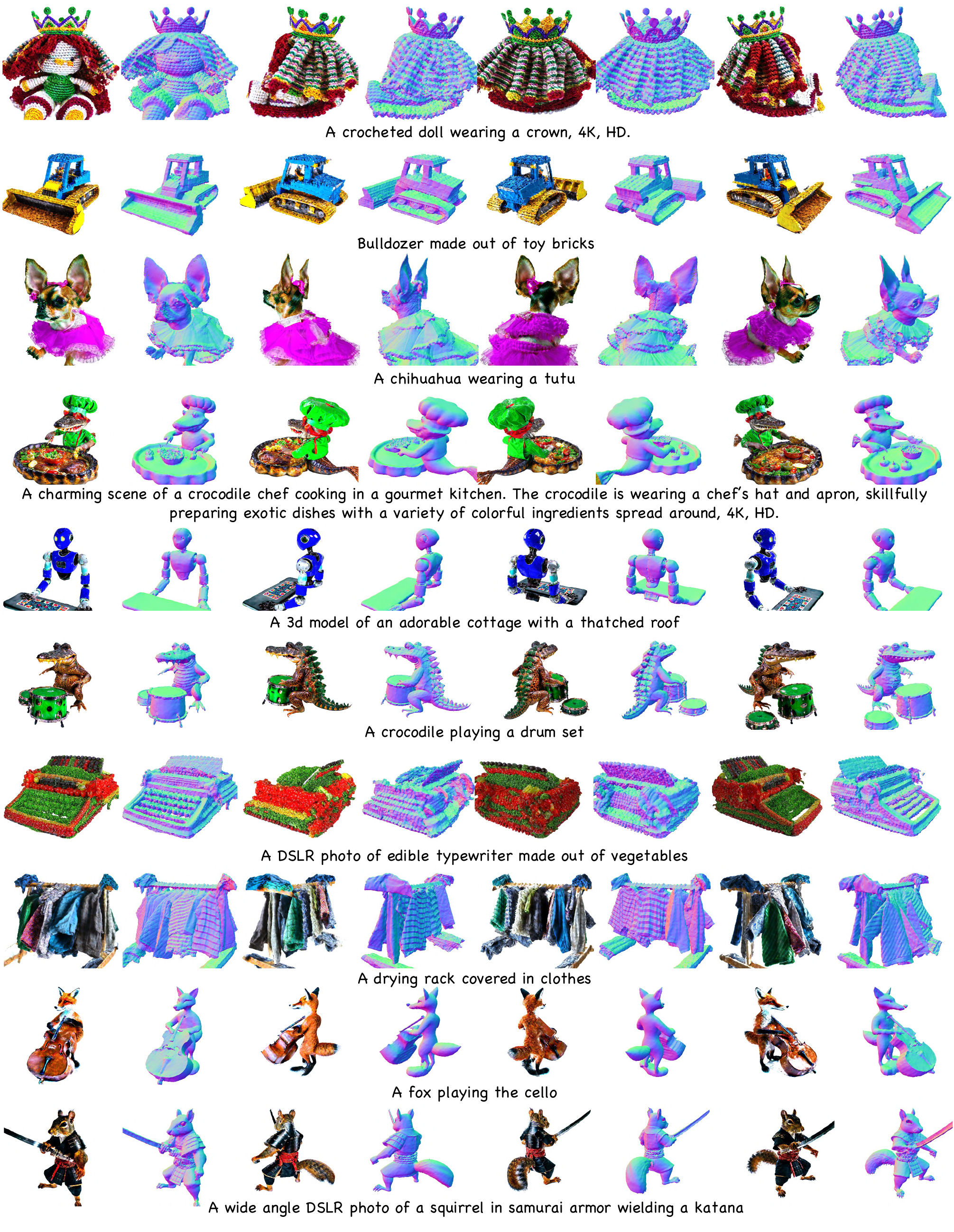}
   \vspace{-2em}
   \caption{Visual results of \emph{Ours~(Sphere)} (Part I).}
   \label{fig:gallery_sphere_1}
\end{figure*}

\begin{figure*}[h] \centering
    \includegraphics[width=\textwidth]{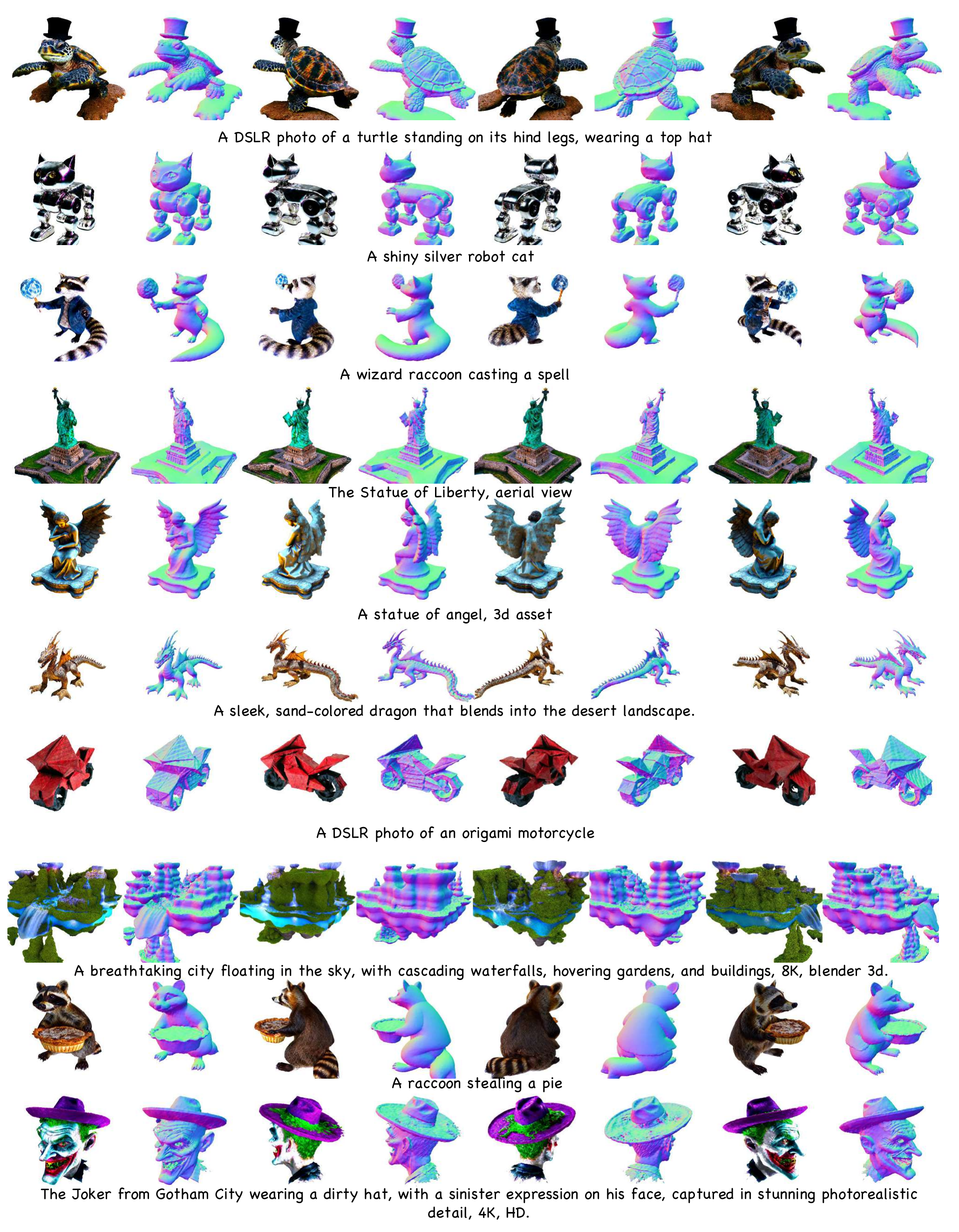}
   \vspace{-2em}
   \caption{Visual results of \emph{Ours~(Sphere)} (Part II).}
   \label{fig:gallery_sphere_2}
\end{figure*}

\begin{figure*}[h] \centering
    \includegraphics[width=\textwidth]{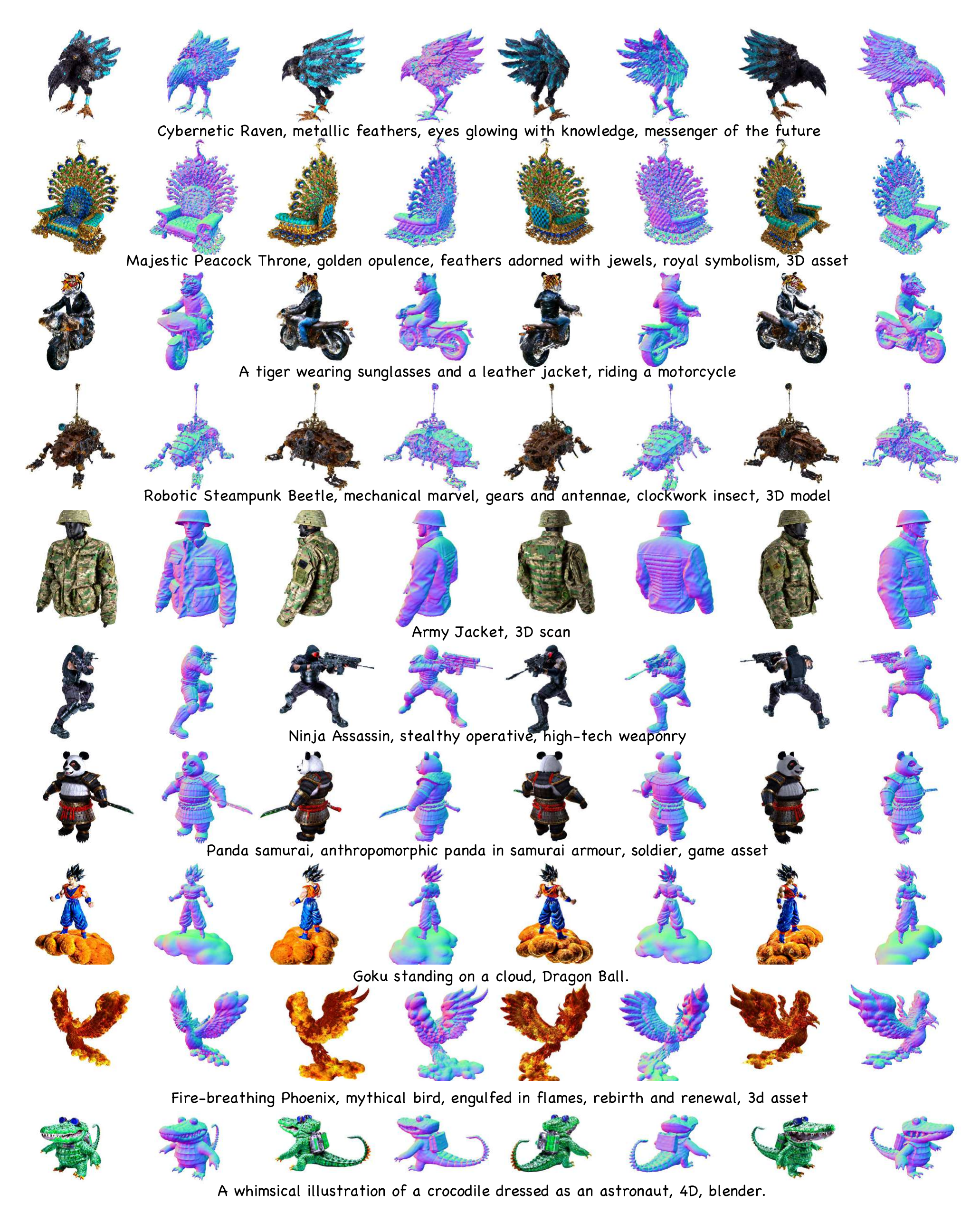}
   \vspace{-2em}
   \caption{Visual results of \emph{Ours~(Sphere)} (Part III).}
   \label{fig:gallery_sphere_3}
\end{figure*}

\begin{figure*}[h] \centering
    \includegraphics[width=\textwidth]{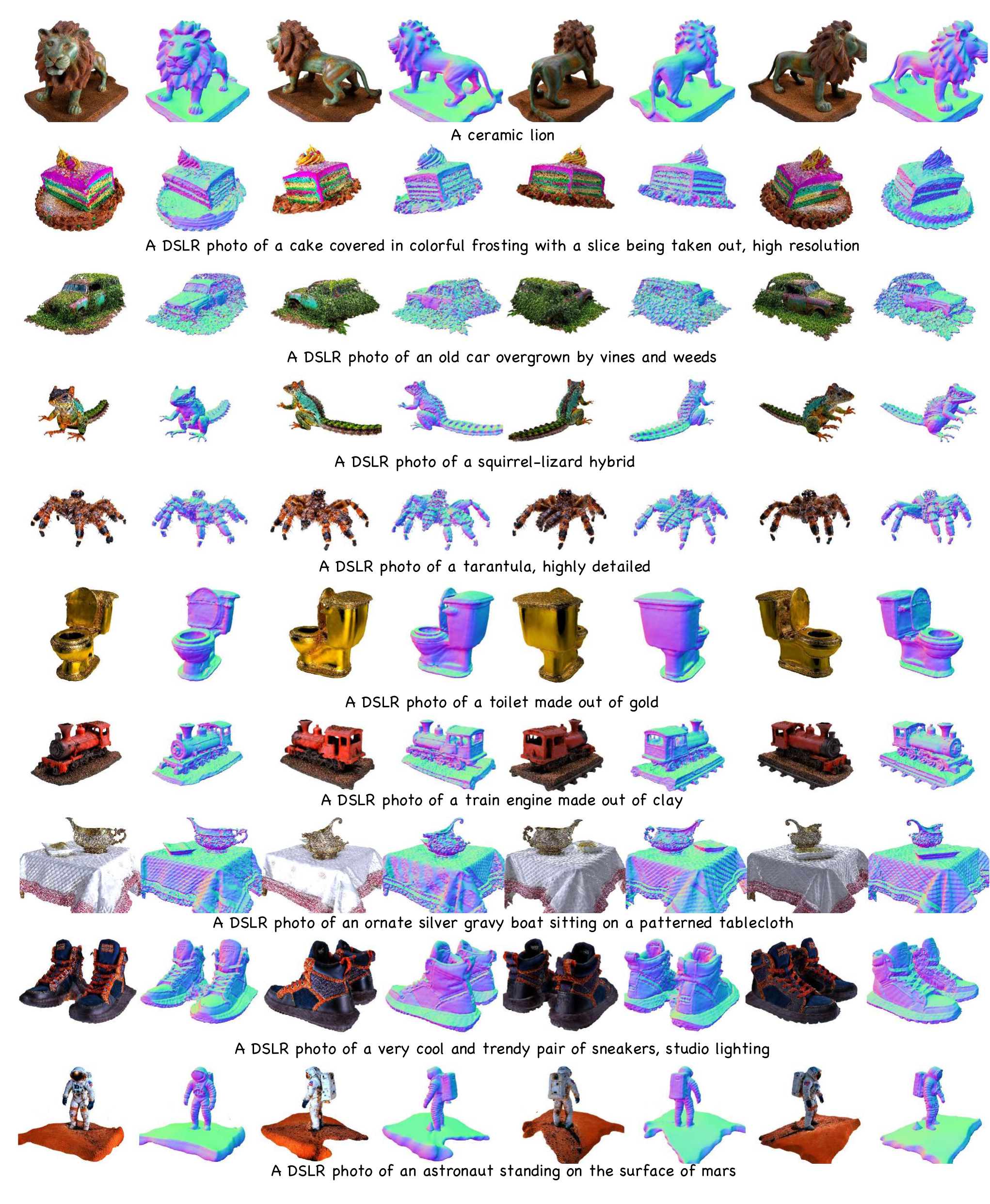}
   \vspace{-2em}
   \caption{Visual results of \emph{Ours~(Sphere)} (Part IV).}
   \label{fig:gallery_sphere_4}
\end{figure*}

\begin{figure*}[h] \centering
    \includegraphics[width=1.0\textwidth]{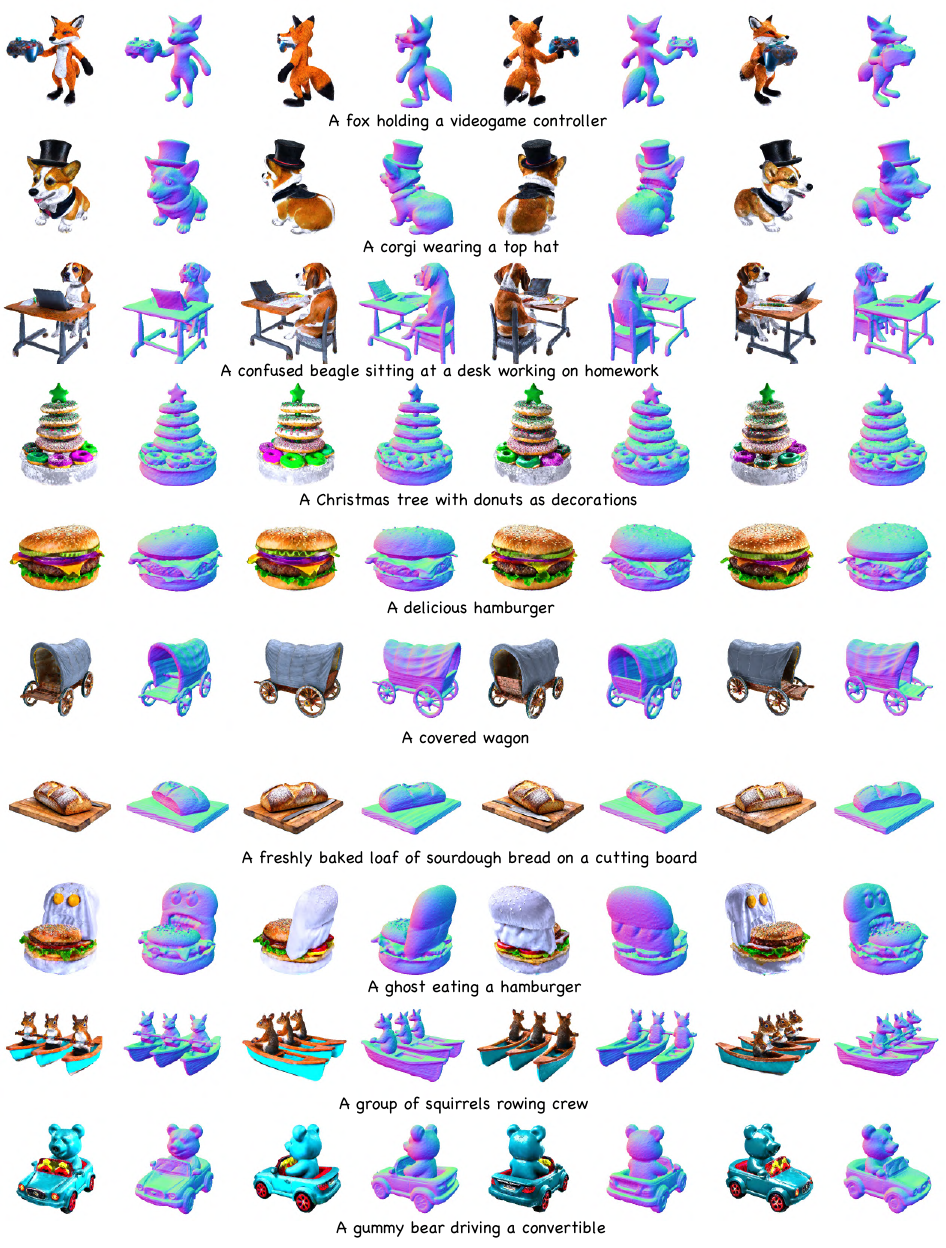}
   \vspace{-7mm}
   \caption{Visual results of \emph{Ours~(NeRF)} (Part I).}
   \label{fig:gallery_nerf_1}
\end{figure*}

\begin{figure*}[h] \centering
    \includegraphics[width=1.0\textwidth]{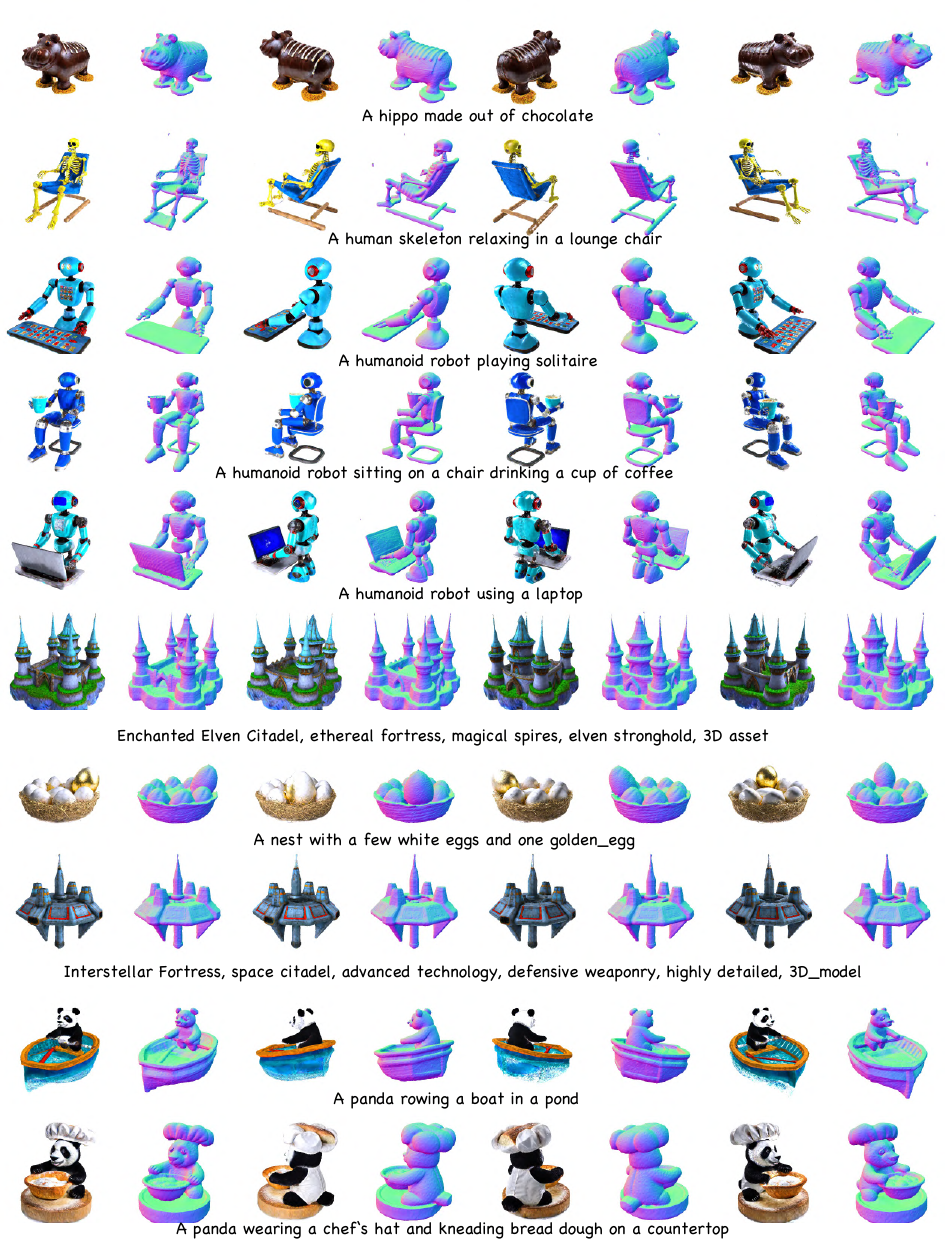}
   \vspace{-7mm}
   \caption{Visual results of \emph{Ours~(NeRF)} (Part II).}
   \label{fig:gallery_nerf_2}
\end{figure*}

\begin{figure*}[h] \centering
    \includegraphics[width=1.0\textwidth]{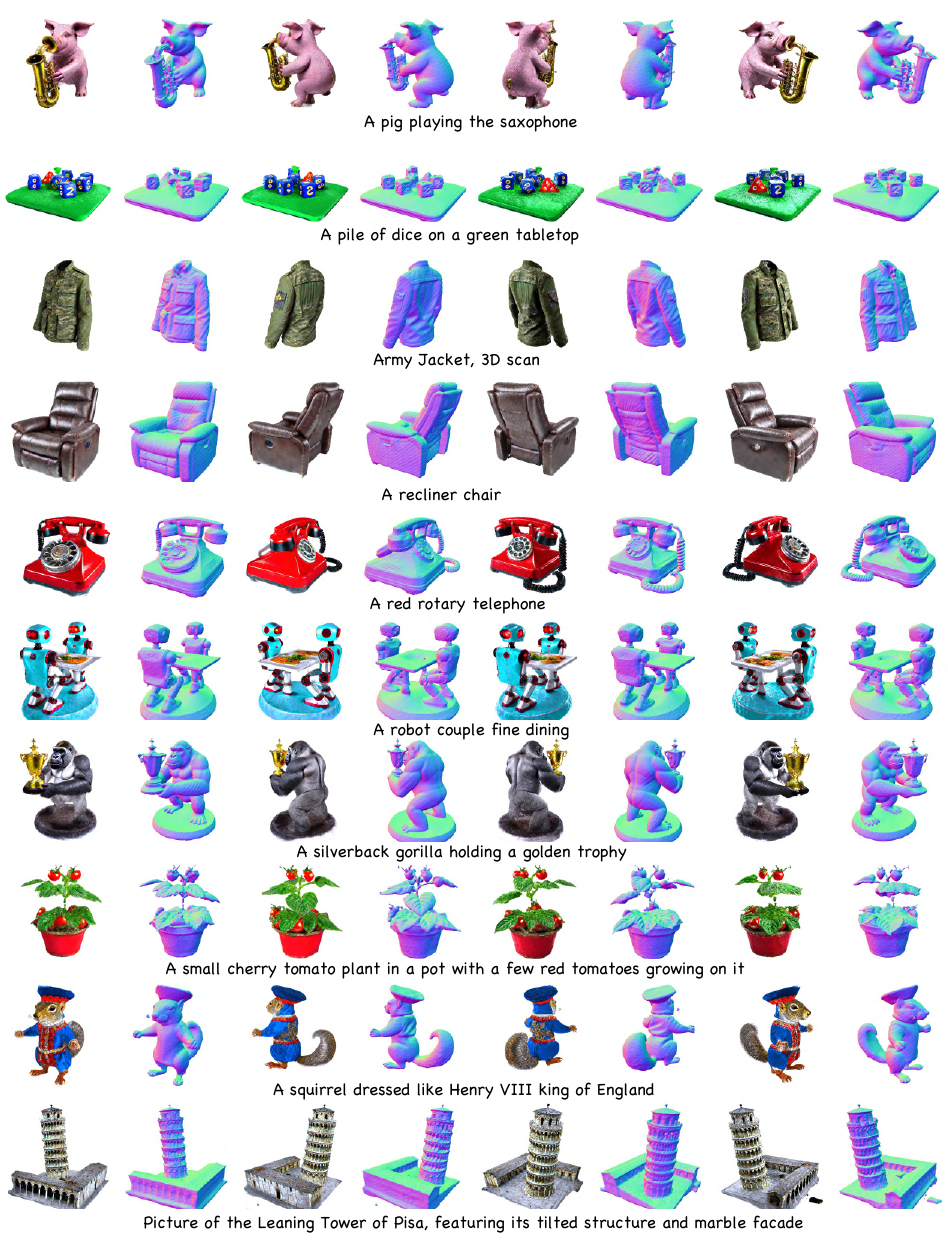}
   \vspace{-7mm}
   \caption{Visual results of \emph{Ours~(NeRF)} (Part III).}
   \label{fig:gallery_nerf_3}
\end{figure*}

\begin{figure*}[h] \centering
    \includegraphics[width=1.0\textwidth]{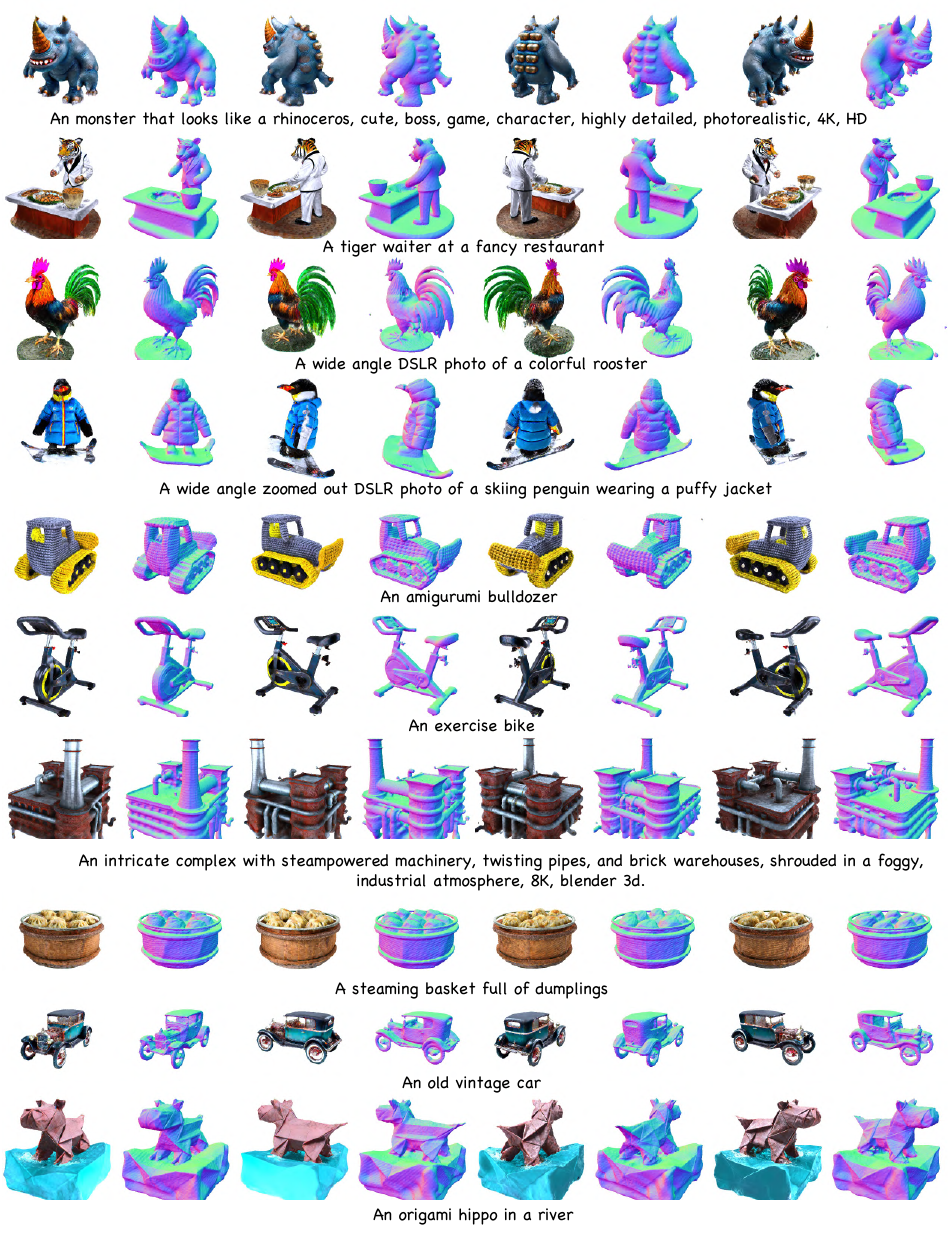}
   \vspace{-7mm}
   \caption{Visual results of \emph{Ours~(NeRF)} (Part IV).}
   \label{fig:gallery_nerf_4}
\end{figure*}

We present more visual results for \emph{Ours~(Sphere)} in Figures~\ref{fig:gallery_sphere_1}- \ref{fig:gallery_sphere_4} and \emph{Ours~(NeRF)} in Figures~\ref{fig:gallery_nerf_1}-\ref{fig:gallery_nerf_4}.

%\xd{More results generated from Our~(NeRF)}

%\todo{The main comparison in Fig. 6 is not fair. All other methods are optimization-based by reusing 2d image diffusion. However, the proposed method used additional datasets. A fair comparison would be zero123+SDS?}

\end{document}